\documentclass[twocolumn]{svjour3}   
\usepackage{amsmath,amssymb}
\usepackage{framed}
\usepackage{url}
\usepackage{psfrag}
\usepackage{graphicx}
\usepackage{subfigure}
\usepackage{multirow}
\usepackage{rotating}
\usepackage{color}
\usepackage{math}
\newcommand{\LineCase}{\textit{Arc Case}}
\newcommand{\CornerCase}{\textit{Corner Case}}
\newcommand{\CircleCase}{\textit{Semi-Spherical Case}}

\newcommand{\correction}[1]{#1}
\journalname{arXiv}
\begin{document} 

\title{
Robust Factorization Methods Using a Gaussian/Uniform\\ Mixture Model
}

\author{Andrei Zaharescu \and Radu Horaud}
\institute{
A. Zaharescu and R. Horaud \at
INRIA Grenoble Rh\^one-Alpes\\
655, avenue de l'Europe \\
38330 Montbonnot, France \\
}

\maketitle
\begin{abstract}
In this paper we address the problem of building a class of robust
factorization algorithms that solve for the shape and motion
parameters with both affine (weak perspective) and perspective camera
models. We introduce a Gaussian/uniform mixture model and its
associated EM algorithm. This allows us to address robust parameter
estimation within a data clustering approach. We propose a robust
technique that works with any affine factorization method and makes it robust
to outliers. In addition, we show how such a framework can be further
embedded into
an iterative perspective factorization scheme. 
We carry out a large number of experiments to validate our algorithms
and to compare them with existing ones.  
We also compare our approach with factorization methods that use M-estimators.

\end{abstract}

\section{Introduction}
\label{section:introduction}

The problem of 3-D reconstruction from multiple images
is central in computer vision \cite{HartleyZisserman00,LuongFaugeras01}. 
Bundle adjustment provides both a general method and practical algorithms for solving this
reconstruction problem
using maximum likelihood
\cite{Triggs:2000}. Nevertheless, bundle adjustment is non-linear in
nature and
sophisticated optimization techniques are necessary, which in turn
require proper initialization.  Moreover, the combination of bundle
adjustment with robust statistical methods to reject outliers is not
clear both from the points of view of convergence properties and of efficiency.
Factorization was introduced by Tomasi \& Kanade \cite{TomasiKanade92} as an elegant
solution to affine multiple-view reconstruction; their initial
solution based on SVD and on a weak-perspective camera model
has
subsequently been improved and elaborated by Morris \& Kanade \cite{Morris:1998}, Anandan
\& Irani \cite{Irani:2000}, Hartley \& Schaffalitzky
\cite{Hartley:2003} as well as by many others. These methods treat the
non-degenarate cases. Kanatani \cite{Kanatani1998},
\cite{Kanatani2004} investigated how to apply model selection
techniques to deal with degenarate cases, namely when the 3-D points lie on a plane
and/or the camera centers lie on a circle.

The problem
can be formulated as the one of minimizing the following Frobenius norm:
\begin{equation}
\vv{\theta}^\ast = \arg \min_{\vv{\theta}} \| \mathbf{S} - \hat{\mathbf{S}}(\vv{\theta}) \|^2_{F}
\label{eq:Frobenius}
\end{equation}
where matrix $\mathbf{S}=[\vv{s}_{ij}]$ denotes the measurement matrix containing matched 2-D image
observations, $\hat{\mathbf{S}}(\vv{\theta}) = \mathbf{M}\mathbf{P}$ denotes the
prediction matrix that can be factorized into the {\em affine} motion matrix
$\mathbf{M}$ and the {\em affine} shape matrix $\mathbf{P}$.  Hence, we
denote by $\vv{\theta}$
the affine motion
{\bf and} shape parameters collectively.
In the error-free
case, direct factorization of the observation matrix using SVD provides an optimal solution. 
More recently
the problem of {\em robust} affine factorization has received a lot of
attention and powerful algorithms that can deal with {\em noisy}, {\em
  missing}, and/or {\em erroneous} data were suggested. 

Anandan \& Irani \cite{Irani:2000} extended the classical SVD approach to deal with the
case of directional uncertainty. They used the Mahalanobis norm instead
of the Frobenius norm and they reformulated the factorization problem
such that the Mahalanobis norm can be transformed into
a Frobenius
norm. This algorithm handles image observations with covariance up to
a few pixels but it cannot cope with missing data, mismatched points,
and/or outliers. More
generally, a central idea is to introduce a weight matrix $\Wmat$ of
the same size as the measurement matrix $\Smat$. The minimization
criterion then
becomes:
\begin{equation}
\vv{\theta}^\ast = \arg \min_{\vv{\theta}} \| \Wmat\otimes(\mathbf{S}-\hat{\mathbf{S}}(\vv{\theta}))\|_{F}^2 
\label{eq:minimization-W}
\end{equation}
where $\otimes$ denotes the Hadamard product ($A = B\otimes C \iff
a_{ij} = b_{ij}c_{ij}$) and $\Wmat=[w_{ij}]$ is
matrix whose entries are weights that
reflect the confidence associated with each image observation.
The most common way of minimizing eq.~(\ref{eq:minimization-W})
is to use alternation methods: these methods are based on the
fact that, if either one of the matrices $\Mmat$ or $\Pmat$ is known, then there is a
closed-form solution for the other matrix that minimizes
eq.~(\ref{eq:minimization-W}). Morris \& Kanade \cite{Morris:1998}
were the first to propose such an alternation method. The PowerFactorization method
introduced by Hartley \& Schaffalitzky \cite{Hartley:2003},
as well as other methods by Vidal \& Hartley  \cite{Vidal:2004}, and
Brant \cite{Brant:2002} fall into this category. PowerFactorization is
based on the PowerMethod for sparse matrix decomposition
\cite{Golub89}. Notice that these techniques are very similar in spirit
with PCA methods with missing data, Wiberg \cite{Wiberg:1976},
Ikeuchi, Shum, \& Reddy
\cite{Shum:1995}, Roweis \cite{Roweis:1997}, and Bishop
\cite{Bishop2006}.  Another way to alternate between the estimation of motion
and shape is to use 
factor analysis, Gruber and Weiss \cite{Gruber:2003}, \cite{Gruber:2004}. 

Alternatively, robustness may be
achieved through {\em adaptive weighting}, i.e., by iteratively updating the weight matrix $\Wmat$ which
amounts to modifying the data $\Smat$, such as is done by
Aanaes et al.
\cite{Aanaes:2002}. 
Their method uses
eq.~(\ref{eq:Frobenius}) in conjunction with a robust loss function
(see Stewart \cite{Stewart:1999uq} and Meer \cite{Meer2004} for details)
to iteratively approximate
eq.~(\ref{eq:minimization-W}), getting a temporary optimum. The
approximation is performed by modifying the original data $\Smat$ such
that the solution to eq.~(\ref{eq:Frobenius}) with {\em modified data}
$\tilde{\Smat}$ is the same as the solution to eq.~(\ref{eq:minimization-W}) with the
original data:
\begin{equation}
\vv{\theta}^\ast = \arg \min_{\vv{\theta}}
\| \Wmat\otimes(\mathbf{S}-\hat{\mathbf{S}}(\vv{\theta})) \|_{F}^2 =
\arg \min_{\vv{\theta}} \| \tilde{\Smat} - \hat{\mathbf{S}}(\vv{\theta})) \|^2_{F}
\label{eq:affine-robust}
\end{equation}
In \cite{Aanaes:2002} the weights are updated via IRLS \cite{Stewart:1999uq}. This
may well be viewed as both an iterative and an alternation
method because the motion matrix
$\Mmat$ is estimated using SVD, than the shape matrix $\Pmat$ is estimated knowing
$\Mmat$, while the image residuals are calculated and the data
(the weights) are modified, etc. 
A similar algorithm that performs outlier correction was proposed by
Huynh, Hartley, \& Heyden \cite{Huynh:2003}. Indeed, if the
observations are noisy, the influence of outliers can be decreased by
iteratively replacing bad observations with ``pseudo''
observations. The convergence of such methods, as
\cite{Aanaes:2002} or \cite{Huynh:2003} is not proved but is tested through
experiments with both simulated and real data. 

An alternative to M-estimators are random sampling techniques
developed independently in computer vision \cite{Fischler:1981fk}
and statistics \cite{Rousseeuw84} (see Meer \cite{Meer2004} for a
recent overview of these methods). For example, Huynh \& Heyden
\cite{Huynh:2002} and Tardif et
al. \cite{Tardif-al2007} use RANSAC, Trajkovic and Hedley use LMedS
\cite{TrajkovicHedley97}, and Hajder and Chetverikov
\cite{HajderChetverikov2004} use LTS (Least Trimmed Squares) \cite{RousseeuwVanAelst99}. The
major drawback of these methods is that they must consider a large
number of subsets sampled from the observation matrix
$\Smat$. 


Generally speaking, robust regression techniques, such as the ones that
we briefly discussed,
work well in conjunction with affine factorization
algorithms. Factorization was initially designed as a ``closed-form solution'' to
multiple-view reconstruction, but robust affine factorization methods are
iterative in nature, as explained above. This has several implications
and some
drawbacks. In the presence of a large number of outliers, proper
initialization is required. The use of an influence function (such as
the truncated quadratic) that tends to zero too quickly cause outliers
to be ignored and hence,  this raises the question of a proper choice
of an influence function. The objective function is non-convex implying that
IRLS will be trapped in local minima. The generalization of affine
factorization to deal with perspective implies the estimation of
depth values associated with each reconstructed point. This is
generally performed iteratively \cite{ST96}, \cite{Christy:1996}, \cite{Mahamud:2000},
\cite{Mahamud:2001}, \cite{MiyagawaArakawa2006}, \cite{OliensisHartley2007}. It is not yet clear
at all
how to combine {\em iterative robust methods} with {\em iterative
projective/perspective factorization methods}. 

In this paper we cast the problem of robust factorization into the
framework of
data clustering \cite{FraleyRaftery2002}. Namely, we consider the problem of
classifying the observed 2-D matched points into two categories:
inliers and outliers. For that purpose we model the likelihood of the
observations with a Gaussian/uniform mixture model. This leads to a
maximum likelihood formulation with missing variables that can be solved
with the EM algorithm \cite{DempsterLairdRubin77},
\cite{McLachlanKrishnan97}, \cite{FraleyRaftery2002}. Notice that this
approach is different than the method proposed by Miller \& Browning
\cite{MillerBrowning2003} requiring both labeled and ulabeled data sets.

We devise an EM algorithm within the framework of 3-D reconstruction
and within the 
specific mixture model just outlined; This immediately
implies convergence of the proposed algorithms, i.e., maximization of the joint likelihood of
the observations. We show that
in this particular case (normally distributed inliers and uniformly
distributed outliers) the posterior propabilities have a very simple
interpretation in terms of robust regression. We describe an affine
factorization algorithm that uses EM; This algorithm is robust and it
shares the convergence properties just outlined. We also describe an
extension of this algorithm to deal with the perspective camera model.

We performed several experiments in two
different scenarios: multiple-camera calibration and 3-D
reconstruction using turn-table data. 
Our method was compared to other
methods on an equal footing: it performs as well as bundle adjustment
to estimate exernal camera parameters. It performs better than IRLS
(used in conjunction with the truncated quadratic) to
eliminate outliers in some difficult cases. 

The remainder of this paper is organized as follows.
Section~\ref{section:probabilistic-modelling} describes the
probabilistic modelling of inliers and outliers using a mixture
between a Gaussian and an uniform distribution. 
Section~\ref{section:EMaffine} explains how this probabilistic model
can be used to derive an affine factorization algorithm and
section~\ref{section:EMperspective} extends this algorithm to
iterative perspective factorization. 
Sections~\ref{section:MultipleCameraCalibration} and
\ref{section:3Dreconstruction} describe experiments performed with
multiple-camera calibration and with multi-view reconstruction data sets.
Section~\ref{section:EM-Mest} compares our approach to M-estimators
and 
section~\ref{section:conclusions} draws some conclusions and gives
some directions for future work.

\section{Probabilistic modelling of inlier/outlier detection}
\label{section:probabilistic-modelling}

The 2-D image points $\svect_{ij}$ ($1\leq i \leq k$, $1\leq j \leq
n$) are the observed values of an equal number of random 
variables $s_{ij}$. We introduce another set of random variables,
$z_{ij}$ which assign a category to each observation. Namely there are
two possible categories, an {\em inlier} category and an {\em outlier}
category. More specifically $z_{ij}=\mbox{inlier}$ means that the
observation $\svect_{ij}$ is an inlier while  $z_{ij}=\mbox{outlier}$
means that the observation $\svect_{ij}$ is an outlier. 

We define the prior probabilities as follows. Let $A_i$ be the area
associated with image $i$ and we assume that all the images have the
same area, $A_i=A$. If an observation is an inlier, then it is
expected to lie within a small circular 
image patch $a$ of radius $\sigma_0$, $a=\pi \sigma_0^2$. The prior
probability of an inlier is the proportion of the image restricted to
such a small circular patch:
\begin{equation}
P(z_{ij}=\mbox{inlier}) = \frac{a}{A}
\label{eq:prior-inlier}
\end{equation}
Similarly, if the observation is an outlier, its probability should
describe the fact that it lies outside this
small patch:
\begin{equation}
P(z_{ij}=\mbox{outlier}) = \frac{A-a}{A}
\label{eq:prior-outlier}
\end{equation}

Moreover, an observation $\svect_{ij}$, given that it is an inlier, should lie in
the neighborhood of an estimation $\hat{\svect}_{ij}$. Therefore, we
will model the probability of an observation  $\svect_{ij}$ {\em given
  that it is assigned to the inlier category} with a Gaussian distribution centered on
$\hat{\svect}_{ij}$ and with a 2$\times$2 covariance matrix $\Cmat$. We obtain:
\begin{align}
\label{eq:likelihood-inlier}
P_{\vv{\theta}}(\svect_{ij} | z_{ij} & =\mbox{inlier})  \\
& = \frac{1}{2\pi(\det\Cmat)^{1/2}}
\exp\left(-\frac{1}{2}d^2(\svect_{ij},\hat{\svect}_{ij}(\vv{\theta}))\right) \nonumber
\end{align}
where we denote by $d$ the Mahalanobis distance:
\begin{equation}
d^2(\svect_{ij},\hat{\svect}_{ij}(\vv{\theta})) = 
(\svect_{ij} -
\hat{\svect}_{ij}(\vv{\theta}))\tp\Cmat\inverse(\svect_{ij} -
\hat{\svect}_{ij}(\vv{\theta}))
\label{eq:distance-observation-estimation}
\end{equation}
Whenever the observation is an outlier, it may lie anywhere in the
image. Therefore, we will model the probability of an observation $\svect_{ij}$ {\em given
  that it is assigned to the outlier category} with a uniform distribution over the image
area:
\begin{equation}
P(\svect_{ij} | z_{ij}=\mbox{outlier}) = \frac{1}{A}
\label{eq:likelihood-outlier}
\end{equation}

Since each variable $z_{ij}$ can take only two values, marginalization
is straightforward and we obtain:
\begin{eqnarray}
P_{\vv{\theta}}(\svect_{ij}) &=& P_{\vv{\theta}}(\svect_{ij} |
z_{ij}=\mbox{inlier}) P(z_{ij}=\mbox{inlier}) \nonumber \\
&+&
P(\svect_{ij} | z_{ij}=\mbox{outlier}) P(z_{ij}=\mbox{outlier})
\nonumber \\
\label{eq:marginal-likelihood}
&=& \frac{a}{2\pi(\det\Cmat)^{1/2}A}
\exp\left(-\frac{1}{2}d^2(\svect_{ij},\hat{\svect}_{ij}(\vv{\theta}))\right) \nonumber \\
&+& \frac{A-a}{A^2}
\end{eqnarray}

We already defined the small area $a$ as a disk of radius $\sigma_0$,
$a=\pi \sigma_0^2$ and we assume that $a \ll A$. 
Using Bayes' formula\footnote{$P(z_{ij}=\mbox{inlier}|\svect_{ij})
  P(\svect_{ij}) = P(\svect_{ij}| z_{ij}=\mbox{inlier}) P(z_{ij}=\mbox{inlier})$ }, 
we
obtain the posterior conditional probability of an observation to be
an inlier. We denote this posterior probability by $\alpha_{ij}^{in}$: 
\begin{align}
\alpha_{ij}^{in} & = P_{\vv{\theta}}( z_{ij}=\mbox{inlier}|
\svect_{ij}) \nonumber \\
& =
\frac{1}{1+\frac{2}{\sigma_0^2} (\det\Cmat)^{1/2}
\exp\left( \frac{1}{2} d^2(\svect_{ij},\hat{\svect}_{ij}(\vv{\theta})) \right)}
\label{eq:inlier-posterior}
\end{align}

\begin{figure*}[h!t!]
\centering
\includegraphics[width=0.7\textwidth]{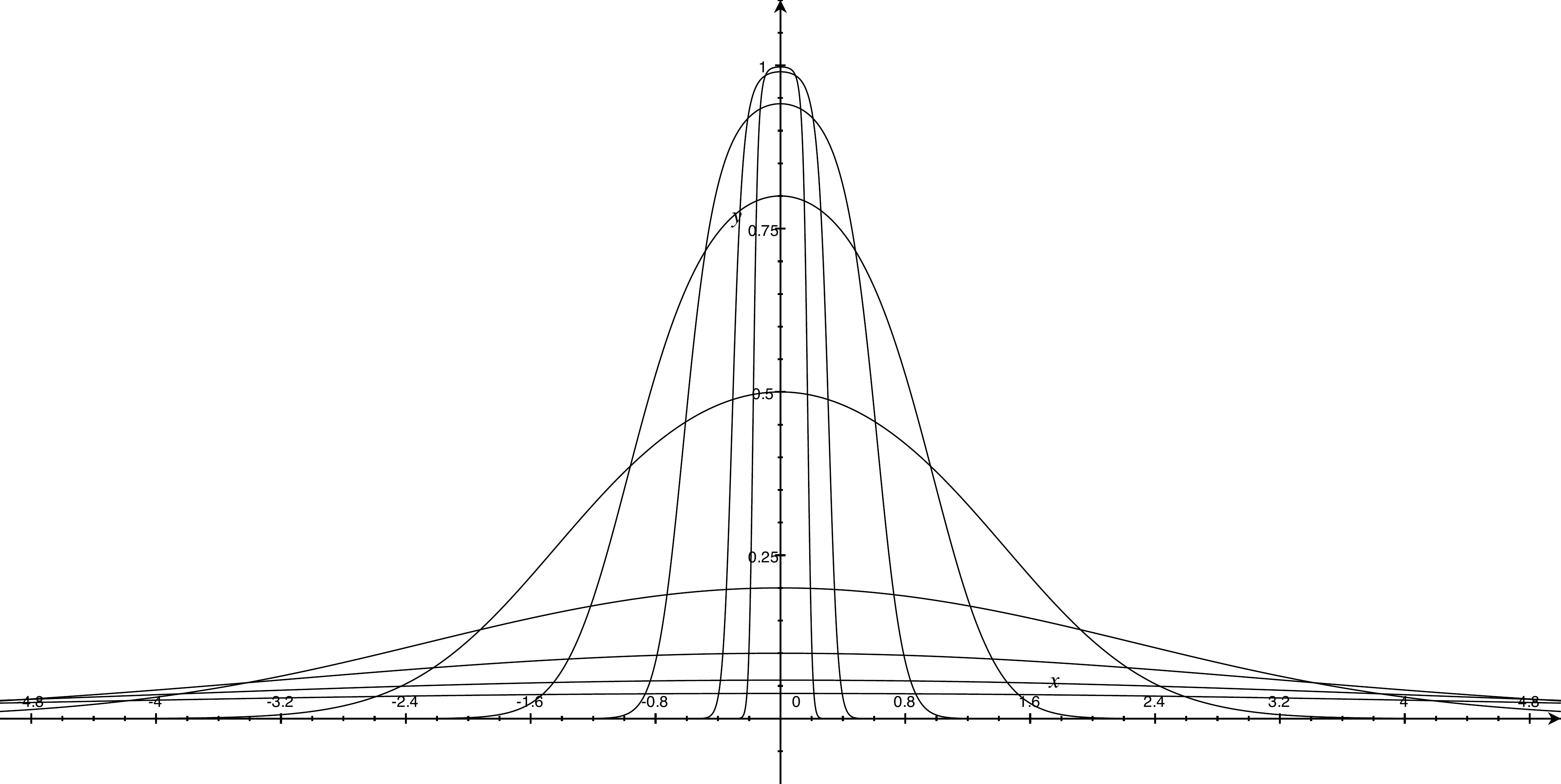}
\caption{Various plots of the conditional posterior probability of an
  observation to be an inlier, i.e.,
  $f_\sigma (x) = 1/(1+\exp(x^2/2\sigma^2))$. This function corresponds to
  eq.~(\ref{eq:inlierposterior-sigma}) with $\sigma_0^2=2$.
As the
  variance decreases, i.e., $\sigma=5,4,3,2,1,0.5,0.25,0.1,0.05$, the function becomes more and more
  discriminant. It is worthwhile to notice that $\lim_{\sigma \to 0}
  f_\sigma (x)$ is a Dirac.
  }
\label{fig:posterior_inlier}
\end{figure*}

The covariance matrix can be written as
$\Cmat=\mm{U}\mm{\Lambda}\mm{U}\tp$ where $\mm{U}$ is a rotation and
$\mm{\Lambda}$ is a diagonal form with entries $\lambda_1$ and
$\lambda_2$. Hence $\det (\Cmat) = \lambda_1\lambda_2$. In order to
plot and illustrate the shape of $\alpha_{ij}^{in}$ as a function of
$\Cmat$ we 
consider the
case of an isotropic covariance, i.e., $\lambda_1=\lambda_2=\sigma^2$
and one may notice that the rotation becomes irrelevant in this
case. We have:
$\Cmat=\sigma^2\Imat_2$. Eq.~(\ref{eq:inlier-posterior}) writes in
this case:
\begin{align}
\alpha_{ij}^{in} & = P_{\vv{\theta}}( z_{ij}=\mbox{inlier}|\svect_{ij}) \nonumber \\
&=
\frac{1}{1+\frac{2\sigma^2}{\sigma_0^2}
\exp\left( \frac{\|\svect_{ij} - \hat{\svect}_{ij}(\vv{\theta})\|^2}{2\sigma^2} \right)}
\label{eq:inlierposterior-sigma}
\end{align}
This posterior probability is shown on Figure~\ref{fig:posterior_inlier}, i.e.,
the function $f_\sigma (x)=1/(1+\sigma^2\exp(x^2/2\sigma^2))$. Here
$\sigma$ takes discrete values in the interval $[0.05,5]$
and 
$\sigma_0^2=2$, i.e., inliers lie within a circle of radius 2 pixels
centered on a prediction. It is worthwhile to notice that, at the limit
$\sigma \rightarrow 0$, we obtain a Dirac function:
\begin{equation}
f_0(x) = \left\{ \begin{array}{cc} 1 & \text{if } x=0\\
                                   0 & \text{if } x\neq 0
\end{array} \right.
\label{eq:Dirac}
\end{equation}

The posterior conditional probability of an observation to be an outlier is given by:
\begin{equation}
\alpha_{ij}^{out}= P_{\vv{\theta}}( z_{ij}=\mbox{outlier}| \svect_{ij} ) = 1 - \alpha_{ij}^{in}
\label{eq:outlier-posterior}
\end{equation}

\subsection{Maximum likelihood with inliers}

The maximum likelihood estimator (ML) maximizes the log-likelihood of
the joint probability of the set of measurements,
$P_{\vv{\theta}}(\Smat)$. Under the assumption that the observations
are independent and identically distributed we have:
\begin{equation}
P_{\vv{\theta}}(\Smat) = \prod_{i,j} P_{\vv{\theta}}(\svect_{ij})
\label{eq:independent-inliers}
\end{equation}

Since we assume that all the observations are inliers, eq.~(\ref{eq:marginal-likelihood})
reduces to:
\begin{equation}
P_{\vv{\theta}}(\svect_{ij})=P_{\vv{\theta}}(\svect_{ij}|z_{ij}=\mbox{inlier})
\label{eq:marginal-inliers}
\end{equation}
The log-likelihood of the joint probability becomes:
\begin{equation}
\log P_{\vv{\theta}}(\Smat) = -\frac{1}{2} \sum_{i,j} \bigg( d^2
(\svect_{ij},\hat{\svect}_{ij}(\vv{\theta})) + \log (\det\Cmat)\bigg)
 + \text{const}
\label{eq:loglikelihood-inliers}
\end{equation}
which can be written as the following criterion:
\begin{align}
\label{eq:minimization-inliers}
&Q_{ML} = \\
& \frac{1}{2}
\sum_{i,j} \bigg( 
(\svect_{ij} -
\hat{\svect}_{ij}(\vv{\theta}))\tp\Cmat\inverse(\svect_{ij} -
\hat{\svect}_{ij}(\vv{\theta})) 
+ \log (\det\Cmat) 
\bigg) \nonumber
\end{align}
The shape and motion parameters can be estimated by minimizing the above
criterion with respect to $\vv{\theta}$:
\begin{equation}
\vv{\theta}^{\ast} = \arg \min_{\vv{\theta}} \frac{1}{2}
\sum_{i,j}
(\svect_{ij} -
\hat{\svect}_{ij}(\vv{\theta}))\tp\Cmat\inverse(\svect_{ij} -
\hat{\svect}_{ij}(\vv{\theta})) 
\label{eq:minimization-ML}
\end{equation}
Once an optimal solution is found, i.e., $\vv{\theta}^{\ast}$, it is
possible to minimize eq.~(\ref{eq:minimization-inliers}) with respect to the
covariance matrix which yields (see appendix \ref{appendix:A}):
\begin{equation}
\Cmat^{\ast} = \frac{1}{m} \sum_{i,j}
(\svect_{ij} -
\hat{\svect}_{ij}(\vv{\theta}^{\ast}))(\svect_{ij} -
\hat{\svect}_{ij}(\vv{\theta}^{\ast}))\tp
\label{eq:covariance-ML}
\end{equation}
where $m=k\times n$ is the total number of observations for $k$ images and
$n$ 3-D points.

Alternatively, if one uses an isotropic covariance, i.e.,
$\Cmat = \sigma^2\Imat$, By minimization of $Q_{ML}$ with respect to $\vv{\theta}$ we
obtain:
\begin{equation}
\vv{\theta}^{\ast} = \arg \min_{\vv{\theta}} \frac{1}{2}
\sum_{i,j}
\| \svect_{ij} -
\hat{\svect}_{ij}(\vv{\theta}))\|^2
\label{eq:minimization-ML-sigma}
\end{equation}
The optimal variance is given by (see appendix \ref{appendix:B}):
\begin{equation}
{\sigma^2}^{\ast} = \frac{1}{2m} \sum_{i,j}
\| \svect_{ij} -
\hat{\svect}_{ij}(\vv{\theta}^{\ast}))\|^2
\label{eq:ML-sigma}
\end{equation}


\subsection{Maximum likelihood with inliers and outliers}

In the presence of outliers, the previous method cannot be
applied. Instead, one has to use the {\em joint probability} of the
observations and of their assignments. Again, by assuming that the
observations are independent, we have:
\begin{eqnarray}
\label{eq:joint-probability-obser-assign}
P_{\vv{\theta}}(\Smat,\Zmat) &=& \prod_{i,j}
P_{\vv{\theta}}(\svect_{ij}, z_{ij}) \\
&=& \prod_{i,j} P_{\vv{\theta}}(\svect_{ij}| z_{ij}) P( z_{ij})
\nonumber \\
&=& \prod_{i,j}
\big( P_{\vv{\theta}}(\svect_{ij}| z_{ij}=\mbox{inlier}) P(
  z_{ij}=\mbox{inlier})
\big)^{\delta_{in}(z_{ij})} \nonumber \\ 
&&
\big( P (\svect_{ij}| z_{ij}=\mbox{outlier}) P(
  z_{ij}=\mbox{outlier})
\big)^{\delta_{out}(z_{ij})} \nonumber 
\end{eqnarray}
The random variables $\delta_{in}(z_{ij})$ and $\delta_{out}(z_{ij})$
are defined by:
\begin{align*}
\begin{array}{cc}
\delta_{in}(z_{ij}) = \left\{
\begin{array}{ll}
1 & \mbox{if } z_{ij}=\mbox{inlier} \\
0 &  \mbox{otherwise} 
\end{array}
\right. 
&
\delta_{out}(z_{ij}) = \left\{
\begin{array}{ll}
1 & \mbox{if } z_{ij}=\mbox{outlier} \\
0 &  \mbox{otherwise} 
\end{array}
\right.
\end{array}
\end{align*}
By taking the logarithm of the above expression and grouping constant terms, we obtain:
\begin{eqnarray}
\log P_{\vv{\theta}}(\Smat,\Zmat)& = &
\sum_{i,j} \big(
\delta_{in}(z_{ij})  \log (P_{\vv{\theta}}(\svect_{ij}|
z_{ij}=\mbox{inlier}) )  \big. \\
\label{eq:log-joint-likelihood}
&+& \big.
\delta_{out}(z_{ij})  \log (P_{\vv{\theta}}(\svect_{ij}| z_{ij}=\mbox{outlier}) ) 
+ \text{const} \big) \nonumber 
\end{eqnarray}
This cannot be solved as previously because of the presence of the
missing
assignment variables $z_{ij}$. Therefore, they will be treated within
an expectation-maximization framework. For this purpose we
evaluate the {\em conditional expectation} of the log-likelihood over the random
variables $z_{ij}$, given the observations $\Smat$:
\begin{align}
\label{eq:expectation-assignments}
E_Z & \left[ \log (P_{\vv{\theta}}(\Smat,\Zmat)) | \Smat \right] \\
&=
\sum_{i,j} \big( \log (P_{\vv{\theta}}(\svect_{ij}| z_{ij}=\mbox{inlier}) )
E_Z \left[ \delta_{in}(z_{ij}) | \Smat \right] \big. \nonumber \\
& +  \big. \log (P(\svect_{ij}| z_{ij}=\mbox{outlier}) ) E_Z \left[
    \delta_{out}(z_{ij}) | \Smat \right] \big) \nonumber
\end{align}

In this formula we omitted the constant terms, i.e., the terms that do not depend on the parameters $\vv{\theta}$ and $\Cmat$.
The subscript $Z$ indicates that the expectation is taken over the
random variable $z$.
From the definition of $\delta_{in}(z_{ij})$ we have:
\begin{eqnarray*}
E[\delta_{in}(z_{ij})] &= &
\delta_{in}(z_{ij}= \mbox{inlier})P(z_{ij}=\mbox{inlier}) \\
&+ &
\delta_{in}(z_{ij}= \mbox{outlier})P(z_{ij}=\mbox{outlier})\\
&=& P(z_{ij}=\mbox{inlier})
\end{eqnarray*}
Hence:
\begin{eqnarray*}
E_Z \left[ \delta_{in}(z_{ij}) | \Smat \right] &=&
P(z_{ij}=\mbox{inlier}| \Smat ) \\
&=& P(z_{ij}=\mbox{inlier}| \svect_{ij} ) =
\alpha_{ij}^{in}
\end{eqnarray*}
and:
\begin{equation*}
E_Z \left[ \delta_{out}(z_{ij}) | \Smat \right] = 1 - \alpha_{ij}^{in}
\end{equation*}
Therefore, after removing constant terms, the conditional expectation becomes:
\begin{align}
\label{eq:conditional-expectation}
E_Z & \left[ \log (P_{\vv{\theta}}(\Smat,\Zmat)) | \Smat \right] \\
&=
-\frac{1}{2} \sum_{i,j} \alpha_{ij}^{in}
\bigg( d^2
(\svect_{ij},\hat{\svect}_{ij}(\vv{\theta})) + \log (\det\Cmat)\bigg) \nonumber
\end{align}
This leads to the following criterion:
\begin{align}
Q_{EM} & =
\frac{1}{2}
\sum_{i,j} \alpha_{ij}^{in} \bigg(
(\svect_{ij} -
\hat{\svect}_{ij}(\vv{\theta}))\tp\Cmat\inverse(\svect_{ij} -
\hat{\svect}_{ij}(\vv{\theta}) \nonumber \\
& + \log (\det\Cmat) 
\bigg)
\label{eq:minimization-inliers-outliers}
\end{align}
\section{Robust affine factorization with the EM algorithm}
\label{section:EMaffine}

In this section we provide the details of how the robust affine factorization
problem can be solved iteratively by maximum likelihood via the
expectation-maximization (EM) algorithm \cite{DempsterLairdRubin77},
and how the observations can be classified into either inliers or outliers by
maximum a-posteriori (MAP).

By inspection of equations~(\ref{eq:minimization-inliers}) and
(\ref{eq:minimization-inliers-outliers}) one may observe that the
latter is a weighted version of the former and hence our formulation
has strong simililarities with M-estimators and their practical
solution, namely iteratively reweighted least-squres (IRLS) \cite{Stewart:1999uq}.
Nevertheless, the {\em weights} $\omega_{ij}=\alpha_{ij}^{in}$ were
obtained using a Bayesian approach: they correspond to the posterior conditional
probabilities of the observations (i.e., given that they are inliers), and
such that the equality $\alpha_{ij}^{in}+\alpha_{ij}^{out}=1$ holds for each
observation. The structure and the shape of these posteriors are
depicted by equations~(\ref{eq:inlier-posterior}) and
(\ref{eq:inlierposterior-sigma}) and shown on
Figure~\ref{fig:posterior_inlier}. These probabilities are functions of
the residual but they are parameterized as well by the 2$\times$2
covariance matrix $\Cmat$ associated with the normal probability distribution of
the observations:
One advantage of our formulation over IRLS is that this covariance is
explicitly taken into consideration and 
estimated within the EM algorithm.

It is worthwhile to remark that the minimization of
eq.~(\ref{eq:minimization-ML}) over the affine shape and motion
parameters, i.e., 
$\vv{\theta}$, can be solved using an affine camera model and a
factorization method such that the ones proposed in the literature
\cite{Aanaes:2002,Hartley:2003}. In practice we use the
PowerFactorization method proposed in \cite{Hartley:2003}. The
minimization of eq.~(\ref{eq:minimization-inliers-outliers}) can be
solved in the same way, provided that estimates for the posterior
probabilities $\alpha_{ij}^{in}$ are available. This can be done by
iterations of the EM algorithm:
\begin{itemize}
\item The {\bf E-step} computes the conditional expectation over the
assignment variables associated with each observation, i.e.,
eq.~(\ref{eq:conditional-expectation}). This requires a current
estimate of both $\vv{\theta}$ and $\Cmat$ from which the
$\alpha_{ij}^{in}$'s are updated.
\item The {\bf M-step} maximizes the conditional expectation or,
equivalently, minimizes eq.~(\ref{eq:minimization-inliers-outliers})
with fixed posterior probabilities. This is analogous, but not
identical, with finding the means 
$\vv{\mu}_{ij}$
and a common covariance $\Cmat$ of $m=k\times n$ Gaussian distributions, with
$\vv{\mu}_{ij}=\hat{\svect}_{ij}(\vv{\theta})$. Nevertheless, the means
$\vv{\mu}=\{\vv{\mu}_{11},\hdots, \vv{\mu}_{kn}\}$ are parameterized
by the global variables $\vv{\theta}$. For this reason, the
minimization problem needs a specific treatment (unlike the classical
mixture of Gaussians approach where the means are independent).
\end{itemize}

Therefore $\min_{\vv{\mu}} Q_{EM}$ in the standard EM method must be
replaced by
$\min_{\vv{\theta}} Q_{EM}$ and it does depend on $\Cmat$ in this case:
\begin{equation}
\vv{\theta}^{\ast} = \arg \min_{\vv{\theta}} \frac{1}{2}
\sum_{i,j}\alpha_{ij}^{in}
(\svect_{ij} -
\hat{\svect}_{ij}(\vv{\theta}))\tp\Cmat\inverse(\svect_{ij} -
\hat{\svect}_{ij}(\vv{\theta})) 
\label{eq:minimization-EM}
\end{equation}
Moreover, the
covariance that minimizes eq.~(\ref{eq:minimization-inliers-outliers})
can be easily derived from \cite{Bishop2006}:
\begin{equation}
\Cmat^{\ast} = \frac{1}{\sum_{i,j}\alpha_{ij}^{in}} \sum_{i,j} \alpha_{ij}^{in} 
(\svect_{ij} -
\hat{\svect}_{ij}(\vv{\theta}^{\ast}))(\svect_{ij} -
\hat{\svect}_{ij}(\vv{\theta}^{\ast}))\tp
\label{eq:covariance-EM}
\end{equation}

In many practical situations it is worthwhile to consider the
case of an {\em isotropic
covariance}, in which case the equations above reduce to:

\begin{equation}
\vv{\theta}^{\ast} = \arg \min_{\vv{\theta}} \frac{1}{2}
\sum_{i,j} \alpha_{ij}^{in}
\| \svect_{ij} -
\hat{\svect}_{ij}(\vv{\theta}))\|^2
\label{eq:minimization-ME-sigma}
\end{equation}
and
\begin{equation}
{\sigma^2}^{\ast} = \frac{1}{2\sum_{i,j}\alpha_{ij}^{in}} \sum_{i,j} \alpha_{ij}^{in}
\| \svect_{ij} -
\hat{\svect}_{ij}(\vv{\theta}^{\ast}))\|^2
\label{eq:ME-sigma}
\end{equation}

This may well be
viewed as a special case of model-based clustering \cite{FraleyRaftery2002}.
It was proved \cite{McLachlanKrishnan97} that EM
guarantees convergence, i.e., that $Q_{EM}^{q+1}<Q_{EM}^{q}$, where
the overscript $q$ denotes the $q^{th}$ iteration, and that this
implies the maximization of the joint probability of the observations:
$P_{\vv{\theta}}(\Smat)^{q+1}>P_{\vv{\theta}}(\Smat)^{q}$.
To conclude, the algorithm can be paraphrased as follows:

\begin{framed}
\begin{description} \item[Affine factorization with EM:] \end{description}
\begin{itemize}
\item[] {\em Initialization:} Use the PowerFactorization method to
  minimize eq.~(\ref{eq:minimization-ML}). This provides initial estimates for $\vv{\theta}$ (the
  affine shape and motion parameters).
Estimate $\Cmat$ (the covariance
  matrix) using eq.~(\ref{eq:covariance-ML}). 
\item[] {\em Iterate} until convergence
\begin{itemize}
\item[] {\em Expectation:} Update the values of $\alpha_{ij}^{in}$ according
  to eq.~(\ref{eq:inlier-posterior}) or eq.~(\ref{eq:inlierposterior-sigma}).
\item[] {\em Maximization:} Minimize $Q_{EM}$ over $\vv{\theta}$
  (affine factorization) using either eq.~(\ref{eq:minimization-EM}) or
  eq.~(\ref{eq:minimization-ME-sigma}). Compute the covariance
  $\Cmat$ with eq.~(\ref{eq:covariance-EM}) or the variance $\sigma^2$
  with eq.~(\ref{eq:ME-sigma}).  
  \end{itemize}
\item[] {\em Maximum a posteriori:} Once the EM iterations terminate, choose in between
  inlier and outlier for each observation, i.e., $\max
  \{\alpha_{ij}^{in};\alpha_{ij}^{out}\}$.
\item[]
\end{itemize}
\end{framed}

The algorithm needs initial estimates for
the shape and motion parameters from which an initial covariance matrix can
be estimated. This
guarantees that, at the start of EM, all the residuals have equal
importance. Nevertheless, ``bad'' observations will have a large
associated residual and, consequently, the covariance is
proportionally large.
As the algorithm proceeds, the covariance adjusts to the
current solution while
the posterior probabilities $\alpha_{ij}^{in}$ become more and
more discriminant as depticted on
Figure~\ref{fig:posterior_inlier}. Eventually, observations associated
with
small residuals will be classified as inliers, and observations with
large residuals will be classified as outliers. 

The overall goal of 3-D reconstruction consists of the
estimation of the shape and motion parameters: As just explained, we
embed affine reconstruction in the
M-step. Therefore, with our algorithm, robustness stays {\em outside}
the factorization method at hand -- is it iterative or not -- and hence one
can plug into EM
any factorization procedure. 

\section{Robust perspective factorization}
\label{section:EMperspective}

In this section we address the problem of 3-D reconstruction using
instrinsically calibrated cameras. Moreover, we consider both the weak-perspective
and the perspective camera models, and we explain how the affine
solution provided by factorization can be upgraded to Euclidean
reconstruction. We describe an
algorithm that combines the EM affine factorization algorithm
described above with an iterative perspective factorization algorithm
\cite{Christy:1996,ZHRL06}. This results in a robust
method for solving the 3-D Euclidean reconstruction problem as well as the
multiple-camera calibration problem. 

An image point $\svect=(x,y)$ is the projection of a 3-D point $\tilde{\Xvect}$:
\begin{equation}
x_{ij} = \frac{\mathbf{r}_i^x\cdot \tilde{\mathbf{X}}_j +
  t_i^x}{\mathbf{r}_i^z\cdot \tilde{\mathbf{X}}_j + t_i^z} =   
\frac{\mathbf{a}_{i}^x\cdot \tilde{\mathbf{X}}_j + b_i^x}{\varepsilon_{ij}+1} 
\label{eq:perspective_x}
\end{equation} 
\begin{equation}
y_{ij} = \frac{\mathbf{r}_i^y\cdot \tilde{\mathbf{X}}_j + t_i^y}{\mathbf{r}_i^z\cdot \tilde{\mathbf{X}}_j + t_i^z} =  
\frac{\mathbf{a}_{i}^y\cdot \tilde{\mathbf{X}}_j + b_i^y}{\varepsilon_{ij}+1} 
\label{eq:perspective_y}
\end{equation}
We introduced the following notations: The rotation matrix
$\Rmat_i\tp=[\mathbf{r}_i^x\;\mathbf{r}_i^y\;\mathbf{r}_i^z]$ and the
translation vector $\tmat_i\tp=(t_i^x\;t_i^y\;t_i^z)$ correspond to the motion parameters and they are also denoted the
external camera parameters. Dividing the above equations with the {\em
  depth} $t_i^z$ we obtain a similar set of scaled equations. 
We have: $\mathbf{a}_i^x =
\mathbf{r}_i^x / t_i^z$, $\mathbf{a}_i^y = \mathbf{r}_i^y / t_i^z$,
$b_i^x =  t_i^x / t_i^z$ and $b_i^y =  t_i^y / t_i^z$.

We denote by $\varepsilon_{ij}$ the {\em perspective distorsion}
parameters, namely the following ratios:
\begin{equation}
\varepsilon_{ij} = \frac{\mathbf{r}_i^z \cdot \tilde{\mathbf{X}}_j}{t_i^z}
\label{eq:epsilon}
\end{equation}

Finally, the perspective equations, i.e.,
eqs.~(\ref{eq:perspective_x}) and (\ref{eq:perspective_y}) can be 
written as:
\begin{equation}
\svect_{ij} (1+ \varepsilon_{ij} ) = \Amat_i \Xvect_j
\label{eq:IPF}
\end{equation}
where $\Xvect=(\tilde{\Xvect},1)$ and $\Amat_i$ denotes the following 2$\times$4 matrix:
\[
\Amat_i=\left[ \begin{array}{cc}
\avect_i^x & b_i^x \\
\avect_i^y & b_i^y
\end{array} \right]
\]

From now on we can replace the parameter vector $\vv{\theta}$ with the
affine shape and motion parameters, namely the point set
$\mathcal{X}=\{\Xvect_1,\ldots \Xvect_j, \ldots \Xvect_k\}$ and the
matrix set
$\mathcal{A}=\{\Amat_1,\ldots \Amat_j, \ldots \Amat_n\}$. Using these
notations, eq.~(\ref{eq:minimization-EM}) can now be written as: 
\begin{equation}
\min_{\mathcal{A},\mathcal{X}} \frac{1}{2}
\sum_{i,j}\alpha_{ij}^{in}
(\svect_{ij}(1+ \varepsilon_{ij}) - \Amat_i \Xvect_j)\tp\Cmat\inverse
(\svect_{ij}(1+ \varepsilon_{ij}) - \Amat_i \Xvect_j)
\label{eq:minimization-RPF}
\end{equation}
which can be solved via the EM affine factorization algorithm with
$\varepsilon_{ij}=0,\forall (i,j)$. A
weak-perspective camera model can then be used for upgrading to Euclidean
reconstruction.

The introduction of the perspective camera model adds non null
perspective-distorsion
parameters $\varepsilon_{ij}$, i.e., eq.~(\ref{eq:epsilon}). 
One fundamental observation is the following: If estimates for the parameters
$\varepsilon_{ij}, \forall i\in[1\ldots k],\forall j\in[1\ldots n]$ are
available, then this
corresponds to a weak-perspective camera model that is closer to the
true perspective model. If the true values of the
perspective-distortion parameters are available, the corresponding
weak-perspective model corresponds exactly to the perspective model. Hence, the problem reduces
to affine factorization followed by Euclidean upgrade. Numerous iterative algorithms
have been suggested in the literature for estimating the
perspective-distortion parameters
associated with each 2-D observation, both with uncalibrated and calibrated
cameras \cite{ST96}, \cite{Christy:1996}, \cite{Mahamud:2000},
\cite{Mahamud:2001}, \cite{MiyagawaArakawa2006}, \cite{OliensisHartley2007} to cite just a few. One possibility is to
perform {\em weak-perspective iterations}. Namely, the algorithm starts with
a {\em zero-distorsion} weak-perspective approximation and then, at
each iteration, it updates the perspective distorsions using 
eq.~(\ref{eq:epsilon}). To conclude, the robust perspective
factorization algorithm can be summarized as follows:

\begin{framed}
\begin{description} \item[Robust perspective factorization:] \end{description}
\begin{itemize}
\item[] {\em Initialization:} Set $\varepsilon_{ij}=0, \forall i\in[1\ldots
  k],\forall j\in[1\ldots n]$.
Use the same initialization step as the
  {\bf affine factorization with EM} algorithm.

\item[] {\em Iterate} until convergence:
\begin{itemize}
\item[] {\em Affine factorization with EM:} Iterate until convergence the
  E- and M-steps of the 
  algorithm described in the previous section.
\item[] {\em Euclidean upgrade:} Recover the rotations, translations,
  and 3-D Euclidean coordinates from the affine shape and affine motion parameters.
\item[] {\em Perspective update:} Estimate new values for the parameters
  $\varepsilon_{ij}, \forall i\in[1\ldots k],\forall j\in[1\ldots n]$. If the current
  depth values are identical with the previously estimated ones, then
  terminate, else iterate.
\end{itemize}
\item[] {\em Maximum a posteriori:} After convergence choose in between
  inlier and outlier for each observation, i.e., $\max
  \{\alpha_{ij}^{in};\alpha_{ij}^{out}\}$.
\item[]
\end{itemize}
\end{framed}

\section{Multiple-camera calibration}
\label{section:MultipleCameraCalibration}

In this section we describe how the solution obtained in the previous
section is used within the context of multiple-camera calibration. As
already described above, we are interested in the estimation of the
external camera parameters, i.e., the alignment between a global
reference frame (or the calibration frame) and the reference frame associated with each one of
the cameras. We assume that the internal camera parameters were
accurately estimated using available software. There are many papers
available that address the problem of internal camera calibration either from
3-D reference objects \cite{Faugeras:1993}, 2-D planar objects
\cite{Zhang:2000}, 1-D objects \cite{Zhang:2004} or self-calibration,
e.g., from point-correspondences  \cite{Luong:1997selfcalibration},
\cite{HartleyZisserman00}, \cite{LuongFaugeras01}.
 
\begin{figure*}[!htpb]
\centering
\subfigure[]{\includegraphics[width=0.4\textwidth]{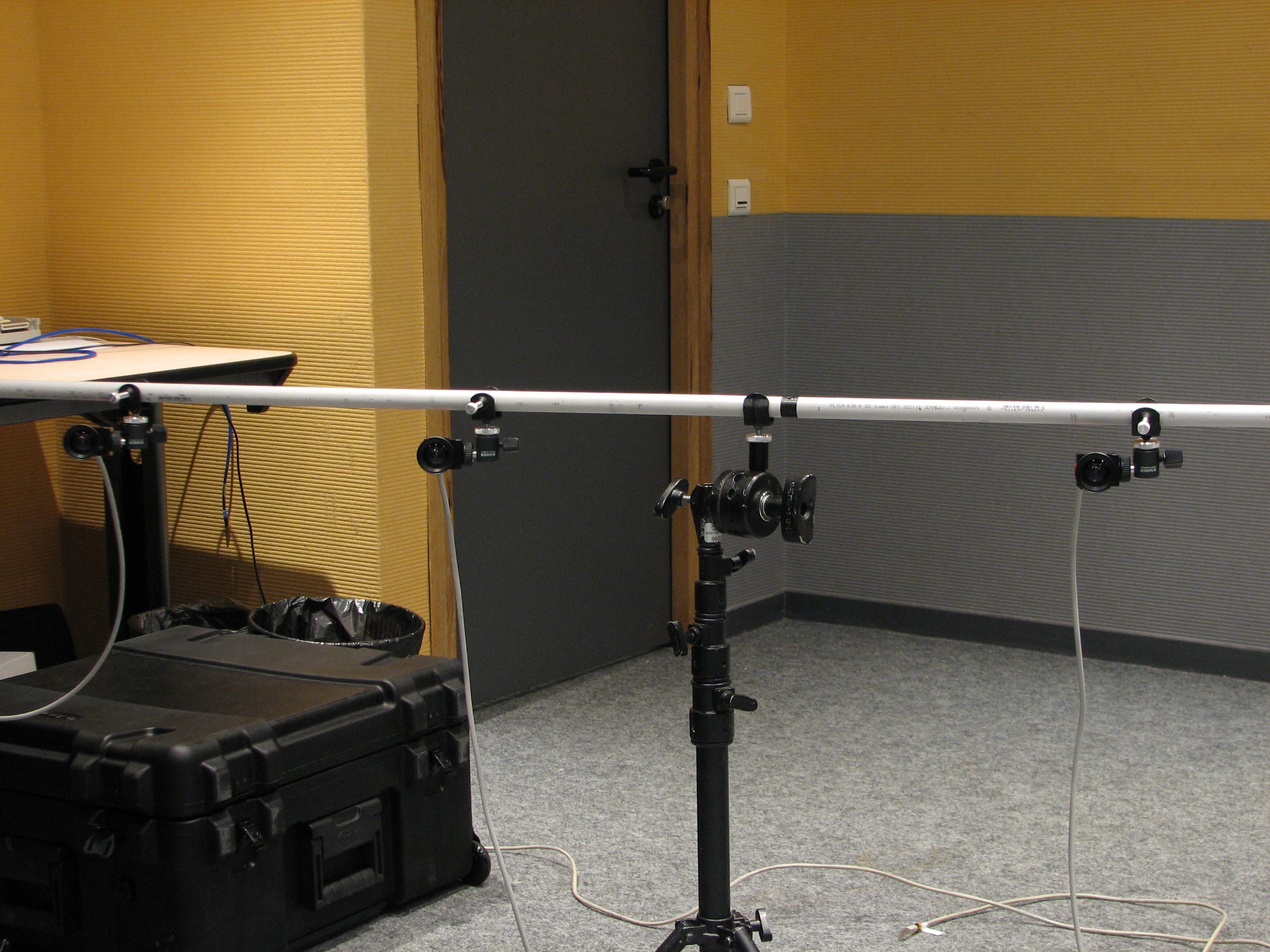}} \quad
\subfigure[]{\includegraphics[width=0.4\textwidth]{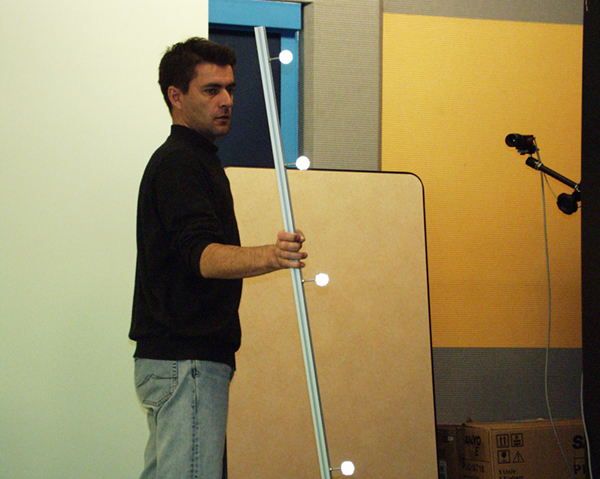}} \quad
\caption{(a): Partial view of a 30-camera setup. (b): The calibration data are gathered by moving a
one-dimensional object in the common field of view of the cameras.}
\label{fig:calib_setup}
\end{figure*}

\begin{figure*}[h!]
\centering
\begin{tabular}{| c | c | c | c | c | c | c | c |}
\hline
 &  \multicolumn{3}{|c|}{Input Image}   & \multicolumn{2}{|c|}{3-D Points}   & \multicolumn{2}{|c|}{Cameras} \\
 \hline
\begin{sideways}\CornerCase\ \end{sideways}  &
\multicolumn{3}{|c|}{ \includegraphics[width=0.26\textwidth]{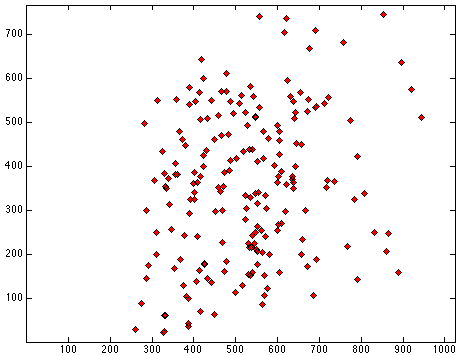} } &
\multicolumn{2}{|c|}{\includegraphics[width=0.26\textwidth]{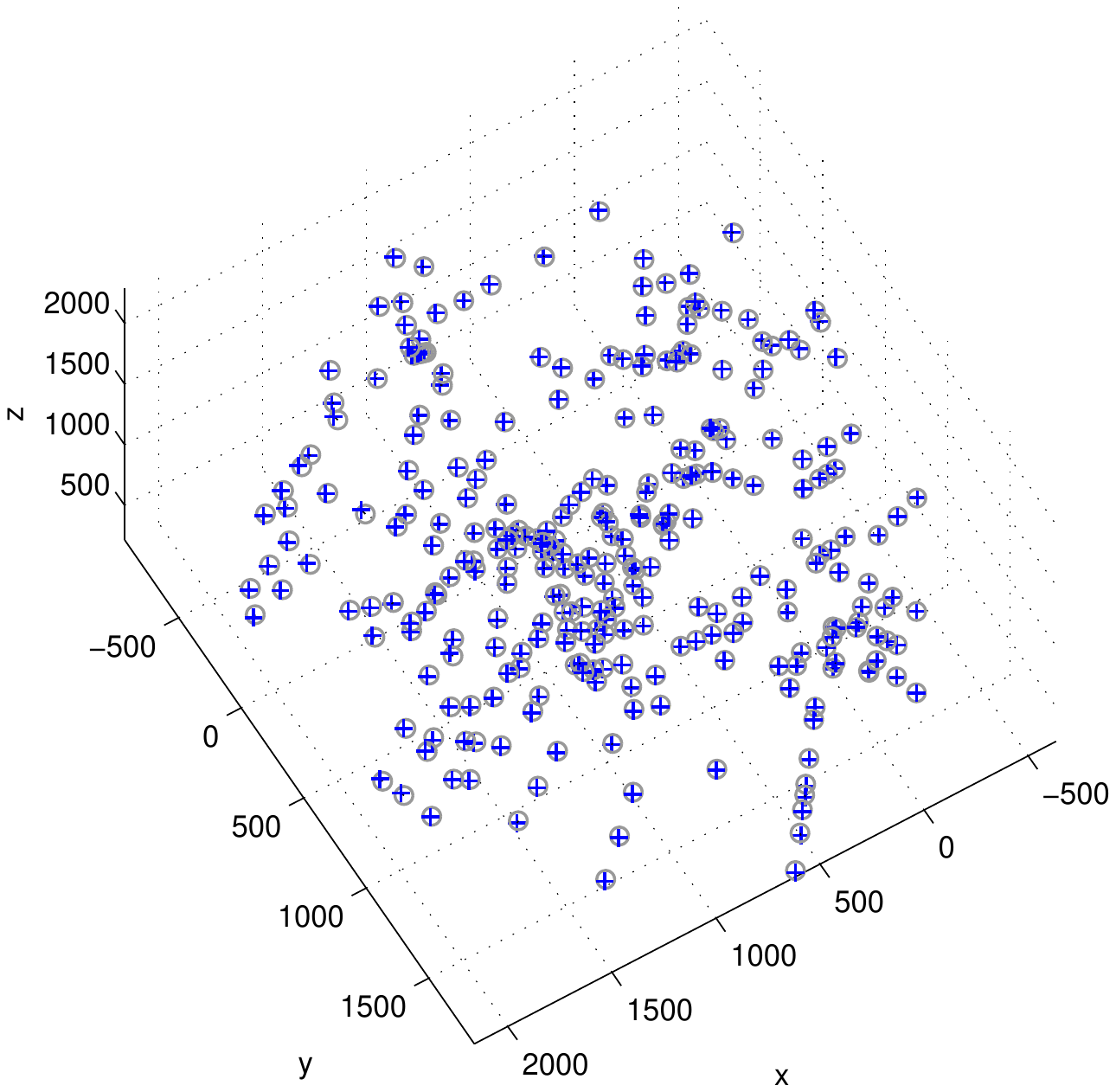}} &
\multicolumn{2}{|c|}{\includegraphics[height=0.26\textwidth]{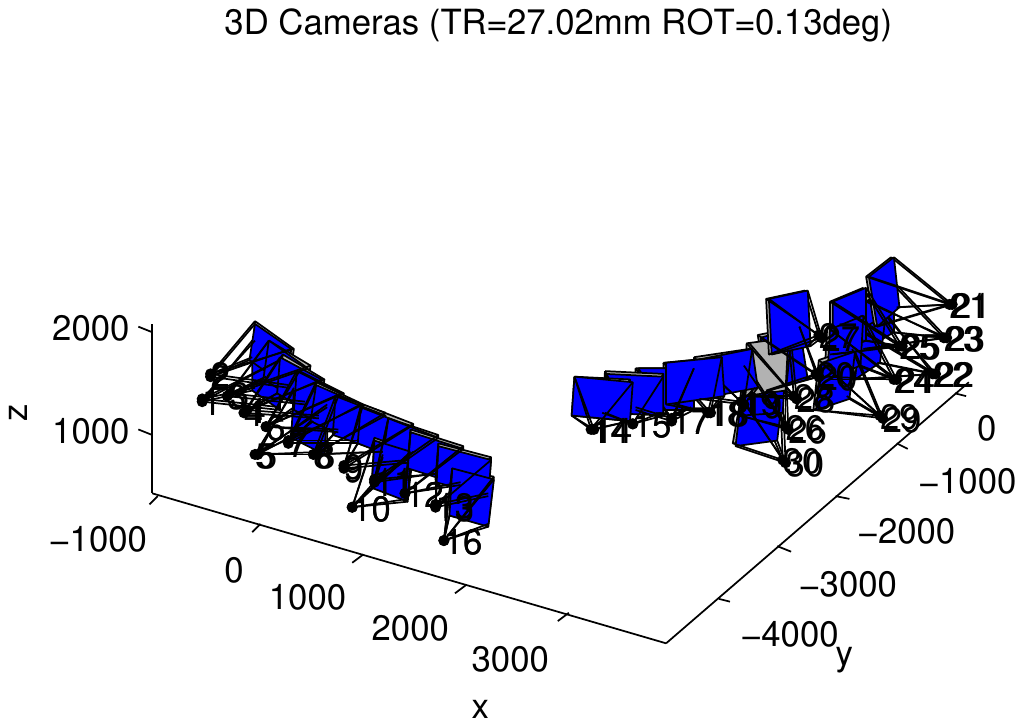}} \\
\hline
\begin{sideways}\LineCase\ \end{sideways}  &
\multicolumn{3}{|c|}{ \includegraphics[width=0.26\textwidth]{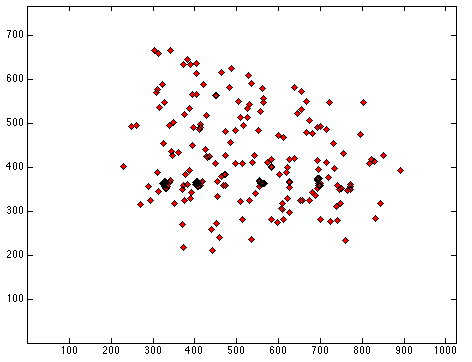} } &
\multicolumn{2}{|c|}{\includegraphics[width=0.26\textwidth]{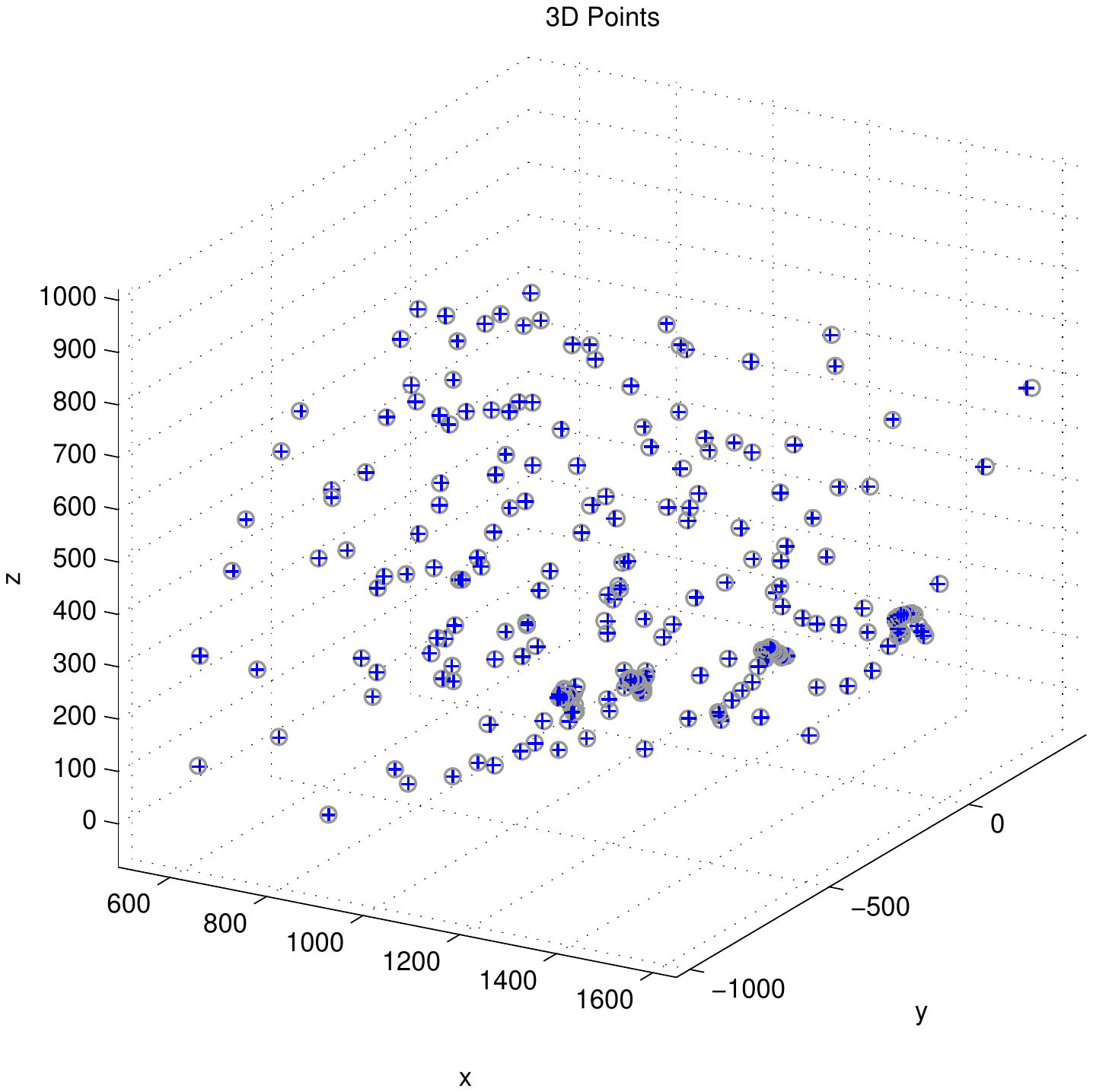}} &
\multicolumn{2}{|c|}{\includegraphics[width=0.26\textwidth]{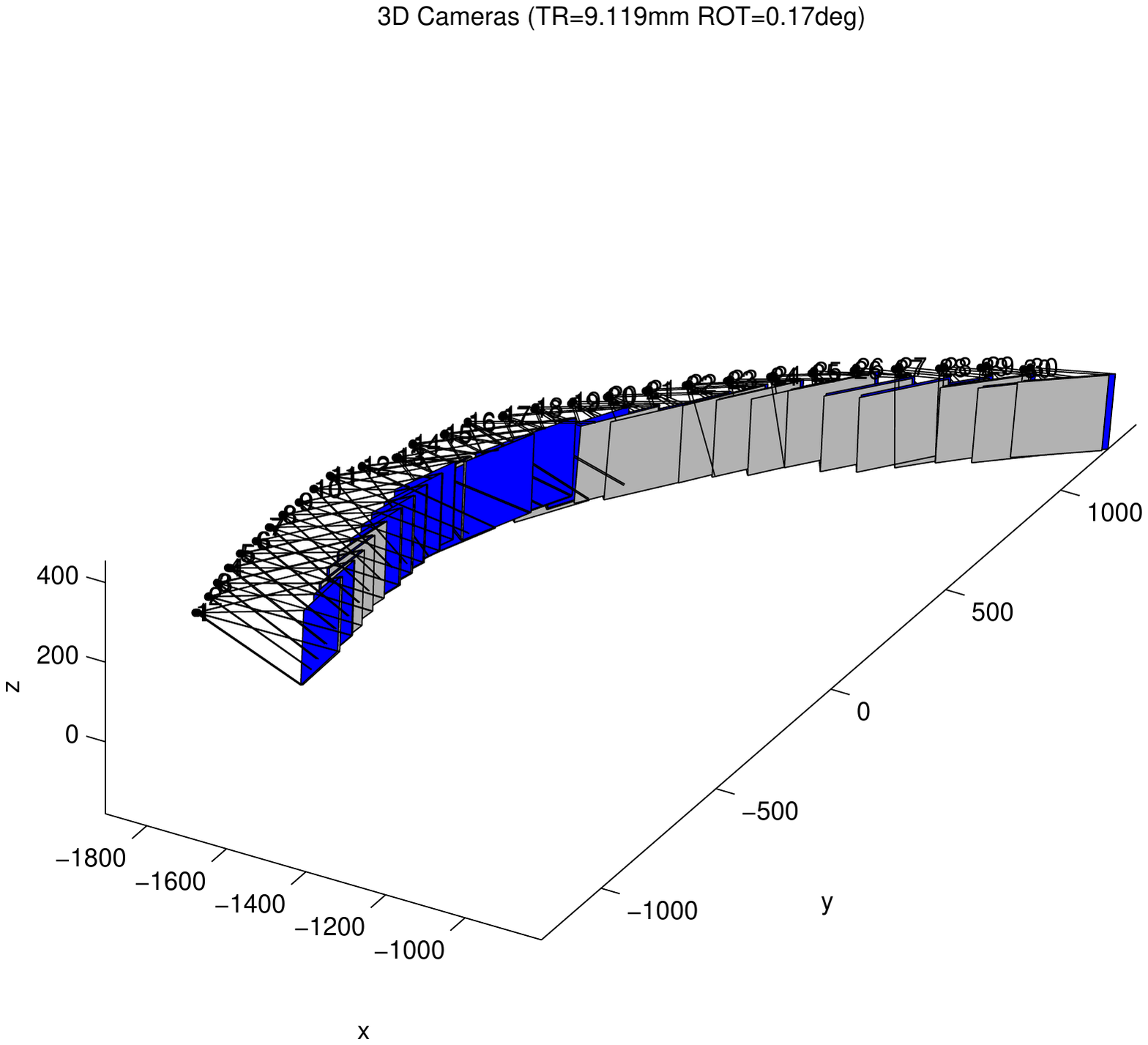}} \\
\hline
\begin{sideways}\CircleCase\ \end{sideways}  &
\multicolumn{3}{|c|}{ \includegraphics[width=0.26\textwidth]{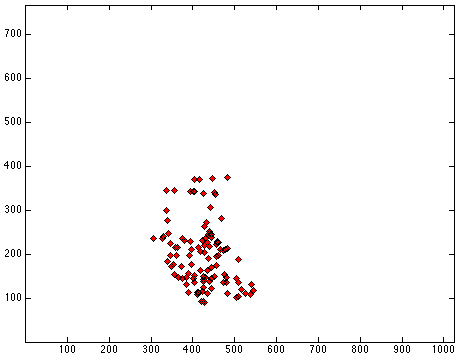} } &
\multicolumn{2}{|c|}{\includegraphics[width=0.26\textwidth]{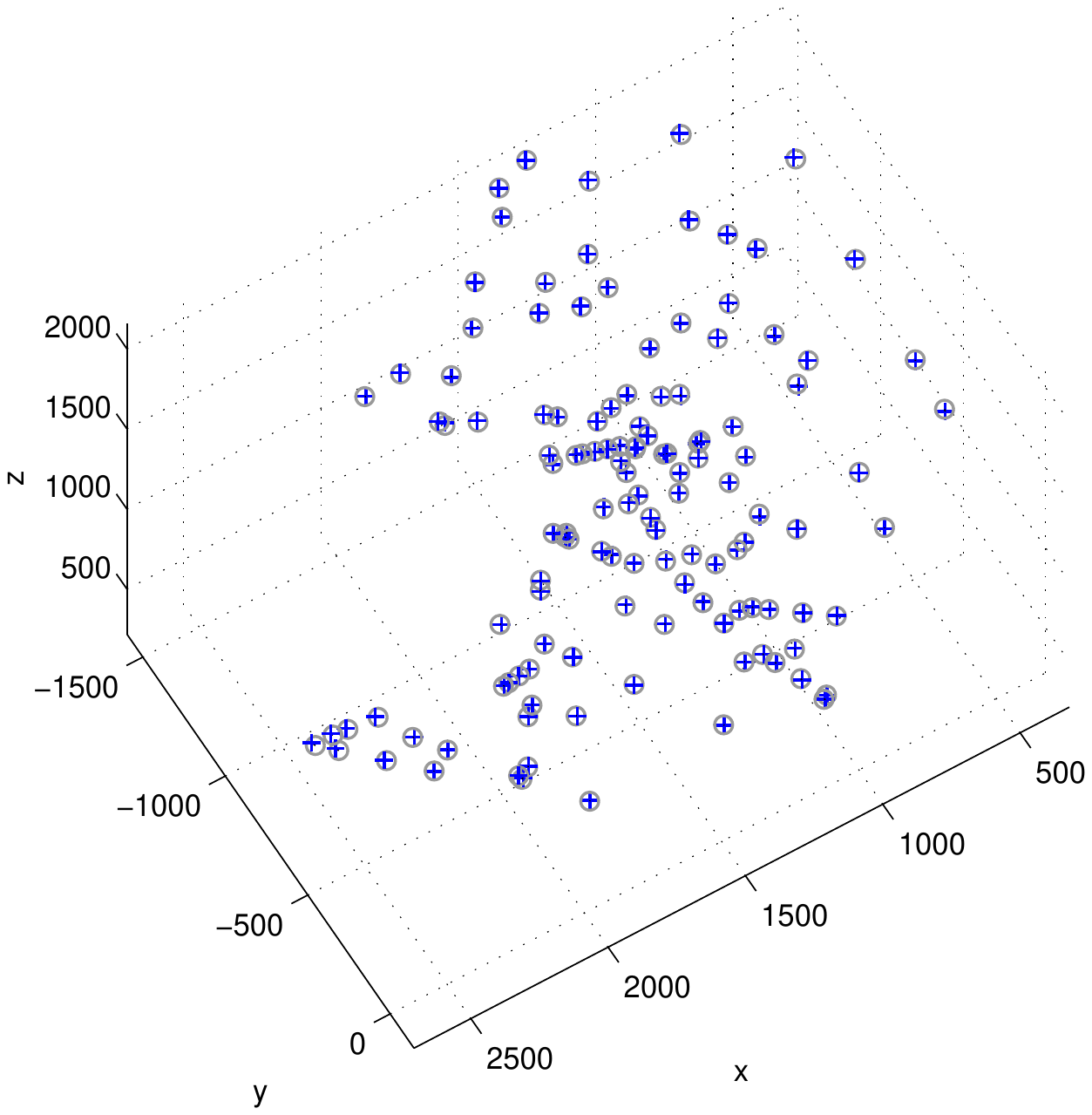}} &
\multicolumn{2}{|c|}{\includegraphics[width=0.26\textwidth]{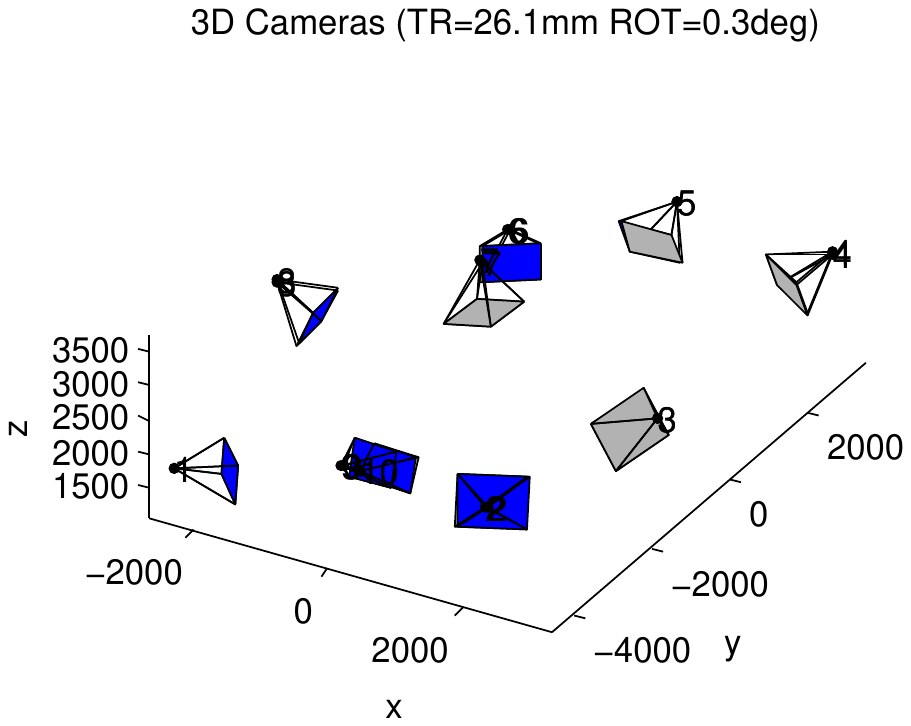}} \\
\hline
\end{tabular}
\caption[calib_results]{Multiple camera calibration results. Left: A typical set of 2-D observations associated with one
  camera. Middle: Reconstructed 3-D points with our method (blue) and
  with bundle adjustment (grey). Right: Camera calibration results
  obtained with our method (blue) and with bundle adjustment (grey) .}
\label{fig:calib_results}  
\end{figure*}

Figure \ref{fig:calib_setup} shows a partial view of a multiple-camera
setup as well as the one-dimensional object used for calibration. In
practice we used three different camera configurations as depicted in
Figure~\ref{fig:calib_results}: two 30 camera configurations and one
10 camera configuration.  These camera setups will be referred to as
the \CornerCase, the \LineCase, and the \CircleCase.
Finding point correspondences accross the images provided
by such a setup is an issue in its own right because one has to solve
for a multiple wide-baseline point correspondence problem. We will briefly
describe the practical solution that we retained and which maximizes
the number of points that are matched over all the views. Nevertheless, in practice
there are missing observations as well as badly detected image
features, bad matches, etc. The problem of missing data has
already been addressed. Here we concentrate on the detection and
rejection of outliers.

We performed multiple camera calibration with two algorithms: The robust perspective
factorization method previously described and bundle
adjustment. We report a detailed comparison between these two
methods. We further compare our robust method with a method based on
M-estimators.

As already mentioned, we use a simple 1-D object composed of four identical markers
with known 1-D coordinates. These coordinates form a
projective-invariant signature (the cross-ratio) that is used to
obtain 3-D to 2-D matches between the markers and their observed image
locations. With finely syncronized cameras it is possible to gather
images of the object while the latter is freely moved in order to cover the 3-D space
that is commonly viewed by all cameras. 
In the three examples below we used
73, 58, and 16 frames, i.e., $292$, $232$,
and $128$ 3-D points. Therefore, in theory there should be $8760$,
$6960$, and $1280$ 2-D observations.

\begin{figure*}[!htb]
\begin{center}
\includegraphics[width=0.9\textwidth]{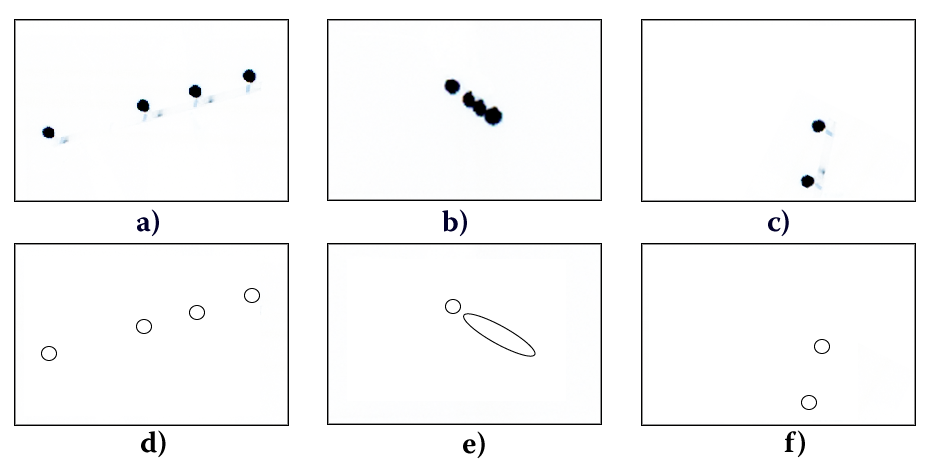}
\caption{Top: These are typical images where the number of connected
  components depend on the position and orientation of the calibrating
  object with respect to the cameras. Bottom: Detected
  blobs with their centers and associated covariance, i.e.,
  second-order moments.}
\label{fig:calibration_frames}
\end{center}
\end{figure*}

Figure \ref{fig:calibration_frames} depicts three
possible image configurations: (a)~four distinct connected components that
correspond without ambiguity to the four markers, (b)~a degenerate view of
the markers, due to strong perspective distorsion, that results in a
number of connected components that cannot be easily matched with the
four markers, and (c)~only two connected components are visible in
which case
one cannot establish a reliable match with the four markers. In
practice we perform a connected-component analysis that finds the number
of blobs in each image. Each such blob is characterized by its center and second
order moments, i.e., Figure \ref{fig:calibration_frames} (d), (e), and
(f). These blobs are matched with the object markers. In
most of the cases the match is unambiguous, but in some cases a blob may be
matched with several markers. 

Let as before, $\svect_{ij}$ denote the center of a blob from image
$i$ that matches marker $j$. The second order moments of this blob
can be used to compute an initial 2$\times$2 covariance matrix
$\Cmat^0_{ij}$ for each such observation. Moreover, we introduce a binary
variable, $\mu_{ij}$, which is equal to 0 if the observation $s_{ij}$
is missing and equal to 1 otherwise.
The multiple-camera calibration algorithm can now be paraphrased as
follows:

\begin{framed}
\begin{description} \item[Multiple camera calibration:] \end{description}
\begin{itemize}
\item[] {\em Initialization:} Use eq.~(\ref{eq:minimization-ML}) to estimate the affine
  shape and motion parameters in the presence of some missing data:
\[
\vv{\theta}^{0} = \arg \min_{\vv{\theta}} \frac{1}{2}
\sum_{i,j} \mu_{ij}
(\svect_{ij} -
\hat{\svect}_{ij}(\vv{\theta}))\tp(\Cmat^0_{ij})\inverse(\svect_{ij} -
\hat{\svect}_{ij}(\vv{\theta})) 
\]
Estimate the initial covariance matrix using eq.~(\ref{eq:covariance-ML}):
\[
\Cmat^{0} = \frac{1}{m} \sum_{i,j} \mu_{ij}
(\svect_{ij} -
\hat{\svect}_{ij}(\vv{\theta}^{0}))(\svect_{ij} -
\hat{\svect}_{ij}(\vv{\theta}^{0}))\tp
\]
Set $\varepsilon_{ij}=0, \forall i\in[1\ldots
  k],\forall j\in[1\ldots n]$
\item[] {\em Iterate} until convergence:
\begin{itemize}
\item[] {\em Affine factorization with EM:} Iterate until convergence the
  E- and M-steps of the 
  algorithm described in section~\ref{section:EMaffine}.
\item[] {\em Euclidean upgrade:} Recover the rotations, translations,
  and 3-D Euclidean coordinates from the affine shape and affine motion parameters.
\item[] {\em Perspective update:} Estimate new values for the parameters
  $\varepsilon_{ij}, \forall i\in[1\ldots k],\forall j\in[1\ldots n]$. If the current
  depth values are identical with the previously estimated ones,
  terminate, else iterate.
\end{itemize}
\item[]
\end{itemize}
\end{framed}

Both the initialization and the M steps of the above algorithm perform
affine factorization in the presence of uncertainty and missing
data. In \cite{ZHRL06} we compared several such algorithms and we came
to the conclusion that the PowerFactorization algorithm outperforms the
other tested algorithms. In order to assess quantitatively the
performance of our algorithm, we compared it with an implementation of
the bundle adjustment method along the lines described in
\cite{HartleyZisserman00}. This comparison requires the estimation of
the rotations and translations allowing the alignment of the two
reconstructed 3-D sets of points with the cameras. We estimate these rotations and
translations using a set of control points. Indeed, both the robust
perspective factorization and the bundle adjustment algorithms need a
number of control points with known Euclidean 3-D coordinates. In practice, the
calibration procedure provides such a set. This set of control points
allows one to
define a global reference frame.
Let $\Pvect_j^c$ denote the 3-D coordinates of the control points
estimated with our algorithm, and let
$\Qvect_j^c$ denote their 3-D coordinates provided in advance. Let
$\lambda$, $\Rmat$, and $\tvect$ be the scale, rotation and
translation allowing the alignment of the two sets of control points. We have:

\begin{equation}
\min_{\lambda,\Rmat,\tvect} \sum_{j=1}^8 \|
\lambda \Rmat \Qvect_j^c + \tvect - \Pvect_j^c \|^{2}
\label{eq:alignment-minimizer}
\end{equation}
The minimizer of this error function can be found in closed form
either with
unit quaternions \cite{Horn87-quat} to represent the rotation $\Rmat$ or with dual-number
quaternions \cite{WalkerShaoVolz92} to represent the rigid motion
$\Rmat,\tvect$. 
Similarly, one can use the same procedure 
to estimate the scale
$\lambda'$, rotation $\Rmat'$, and translation $\tvect'$ associated
with the 3-D reconstruction obtained by bundle adjustment.

Finally, in order to evaluate the quality of the results we estimated the
following measurements:
\begin{description}
\item[The 2D error] is measured in pixels and corresponds
  to the RMS error between the observations and the predictions
  weighted by their posterior probabilities:
\[
\left( \frac{ \sum_{i,j}\alpha^{in}_{ij}\|
  \svect_{ij}-\hat{\svect}_{ij}(\vv{\theta}^{\ast})\|^2}{\sum_{i,j}\alpha^{in}_{ij}}\right)^{1/2}
\]
\item[The 3D error] is measured in milimeters and corresponds to the RMS error between the two sets of 3-D
  points obtained with our algorithm and with bundle adjustment:
\[ \left( \frac{\sum_{j=1}^{n} \| \Pvect_j - \Pvect'_j \|^2}{n}
\right)^{1/2}
\]
\item[The error in rotation] is measured in degrees and depicts the
  average angular error of the 
  rotation matrices over all the cameras. With the same notations as before,
  let $\Rmat_i$ and $\Rmat'_i$ be the rotations of camera $i$ as
  obtained with our algorithm and with bundle adjustment. Let $\vvect$
  be an arbitrary 3-D vector. The dot product $(\Rmat_i\vvect)\cdot
  (\Rmat'_i\vvect)$ is a reliable measure of the cosine of the angular
  discrepancy between the
  two estimations.
Therefore the RMS error in
  rotation can be measured by the angle:
\[
\arccos \left( \frac{180}{\pi} \sqrt{ \frac{\vvect\tp \; \big( \sum_{i=1}^{k} \Rmat_i\tp\Rmat'_i \big)
    \; \vvect}{k}} \right)
\]
\item[The error in translation] is measured in milimeters with:
\[ \left( \frac{\sum_{i=1}^{k} \| \tvect_i - \tvect'_i \|^2}{k}
\right)^{1/2}
\]
\end{description}

As already mentioned, we used three camera setups. All setups use
identical 1024$\times$768 Flea cameras from Point Grey Research
Inc.\footnote{\url{http://www.ptgrey.com/products/flea/}} The intrinsic
parameters were estimated in advance. Two of the setups use 30 cameras,
whereas the third one uses 10 cameras. We denote these setups as
\CornerCase,  \LineCase\,  and  \CircleCase, based on the camera
layout.  

\begin{figure*}[!htb]
\centering
\begin{tabular}{| c | c | c |}
\hline
$1^{st}$ iteration & $2^{nd}$  iteration & $3^{rd}$  iteration \\
\hline
\includegraphics[width=0.30\textwidth]{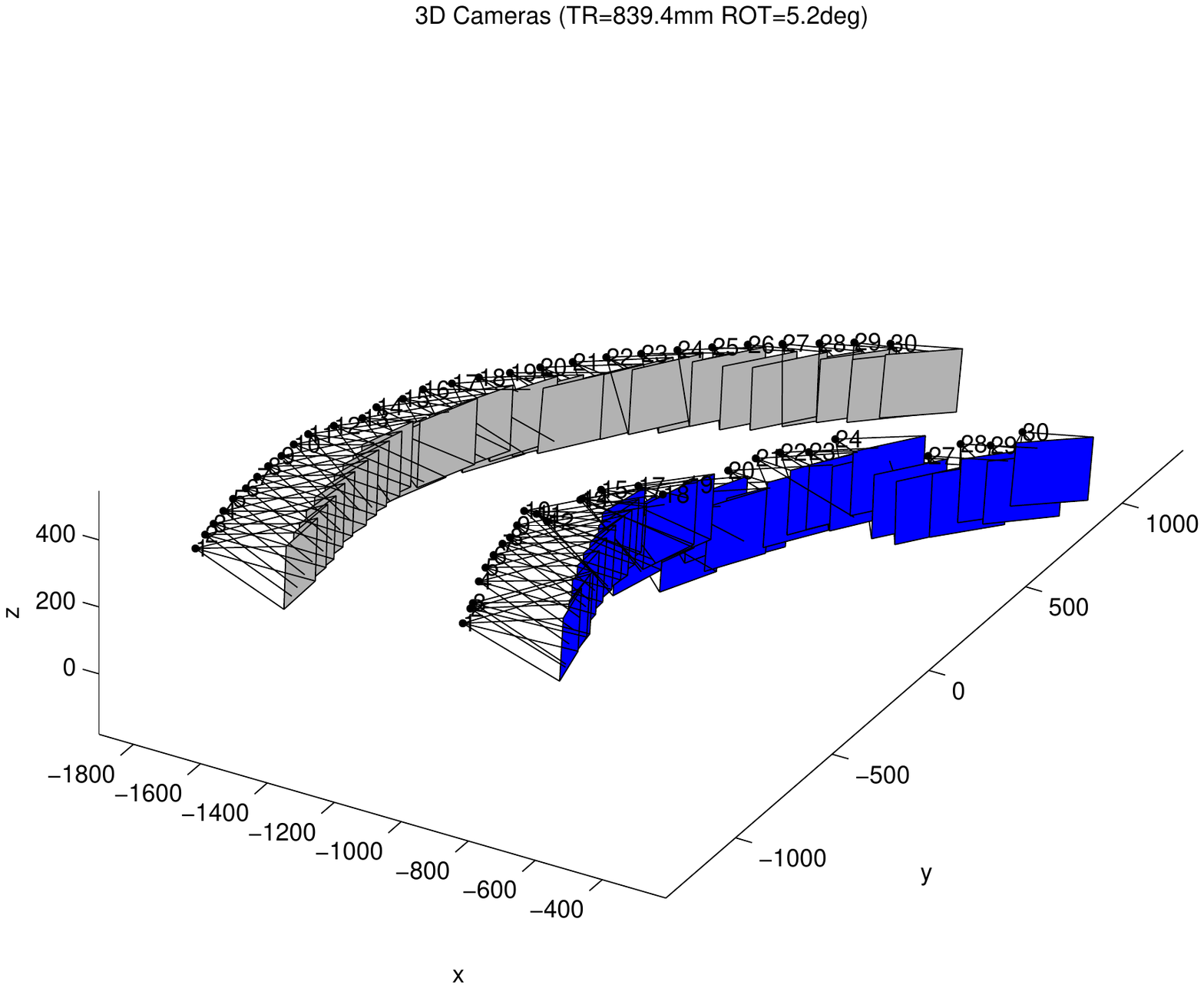} &
\includegraphics[width=0.30\textwidth]{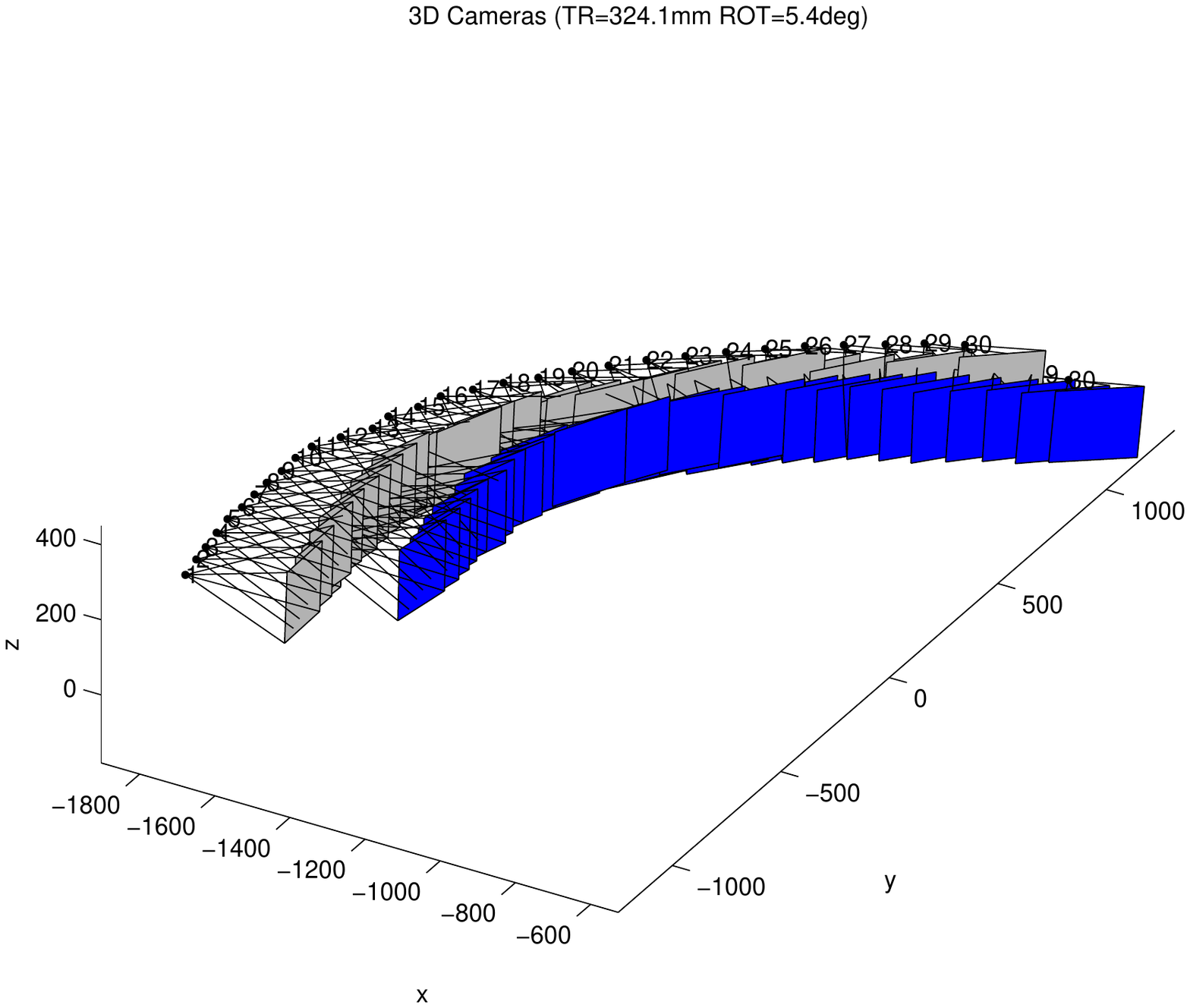} &
\includegraphics[width=0.30\textwidth]{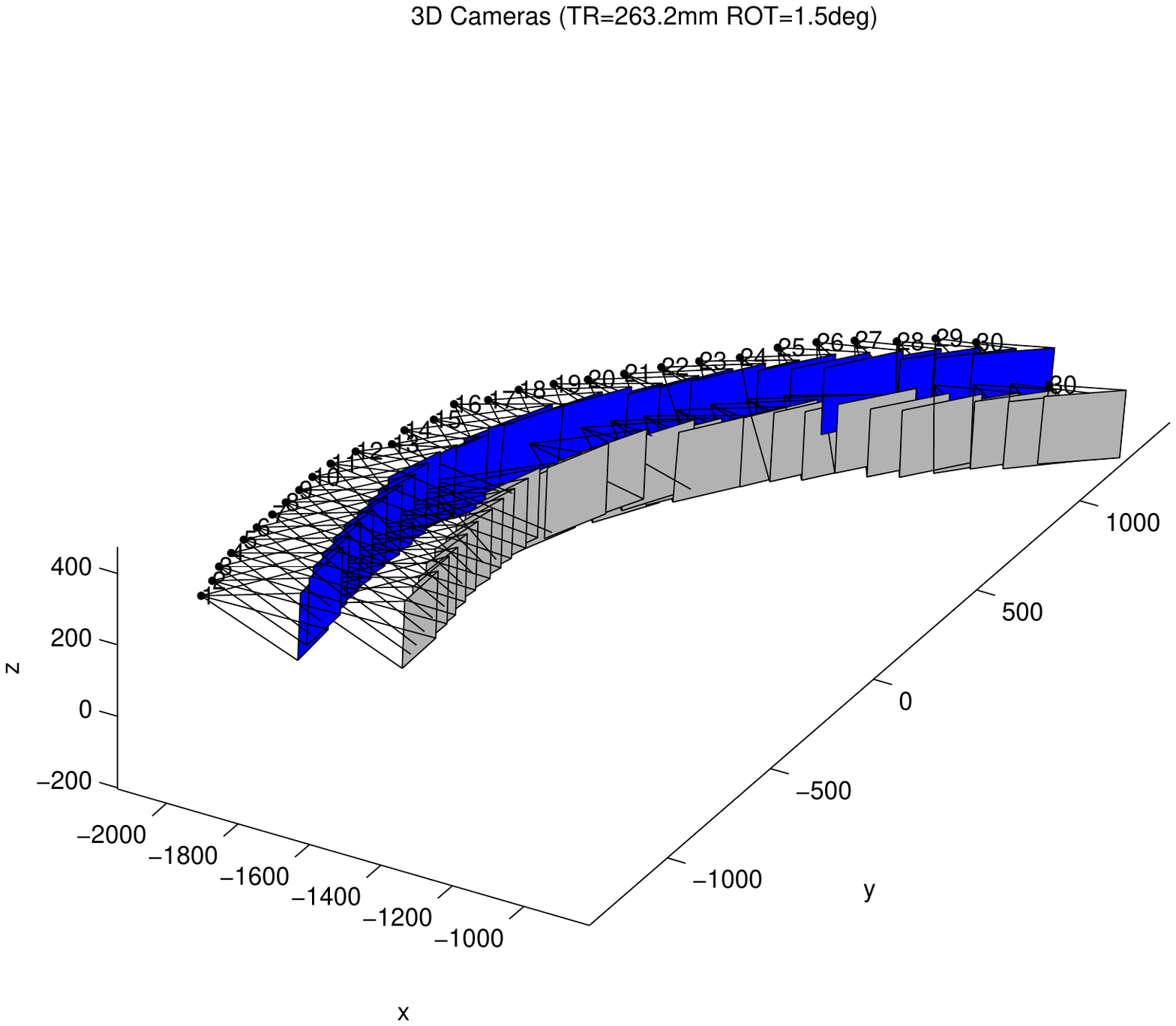}\\
\hline
\small 2D Err. = 59.32 & \small 2D Err.= 18.43 & \small 2D Err. =15.92 \\
\hline
$4^{th}$  iteration & $5^{th}$ iteration & $8^{th}$ iteration \\
\hline
\includegraphics[width=0.30\textwidth]{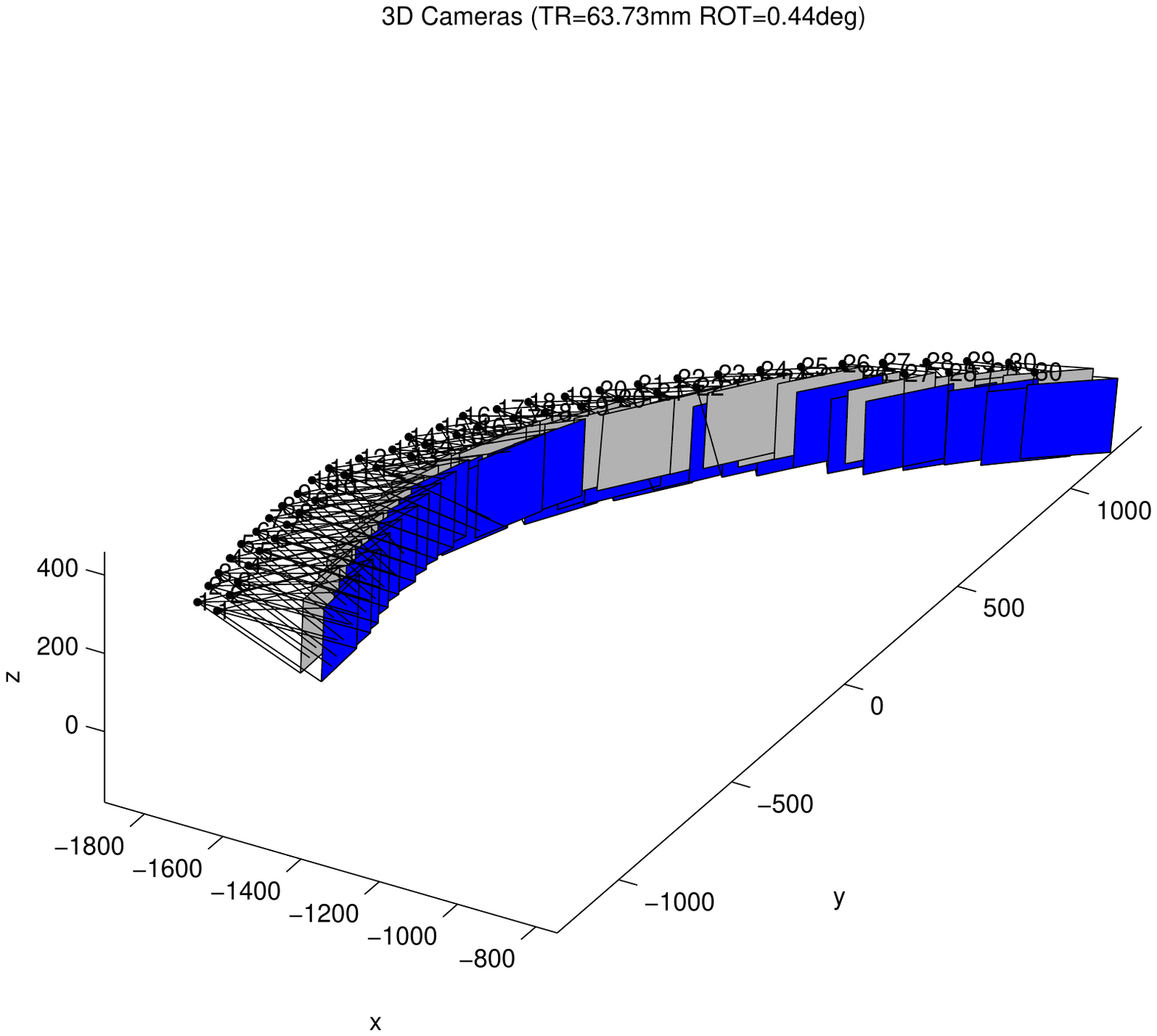} &
\includegraphics[width=0.30\textwidth]{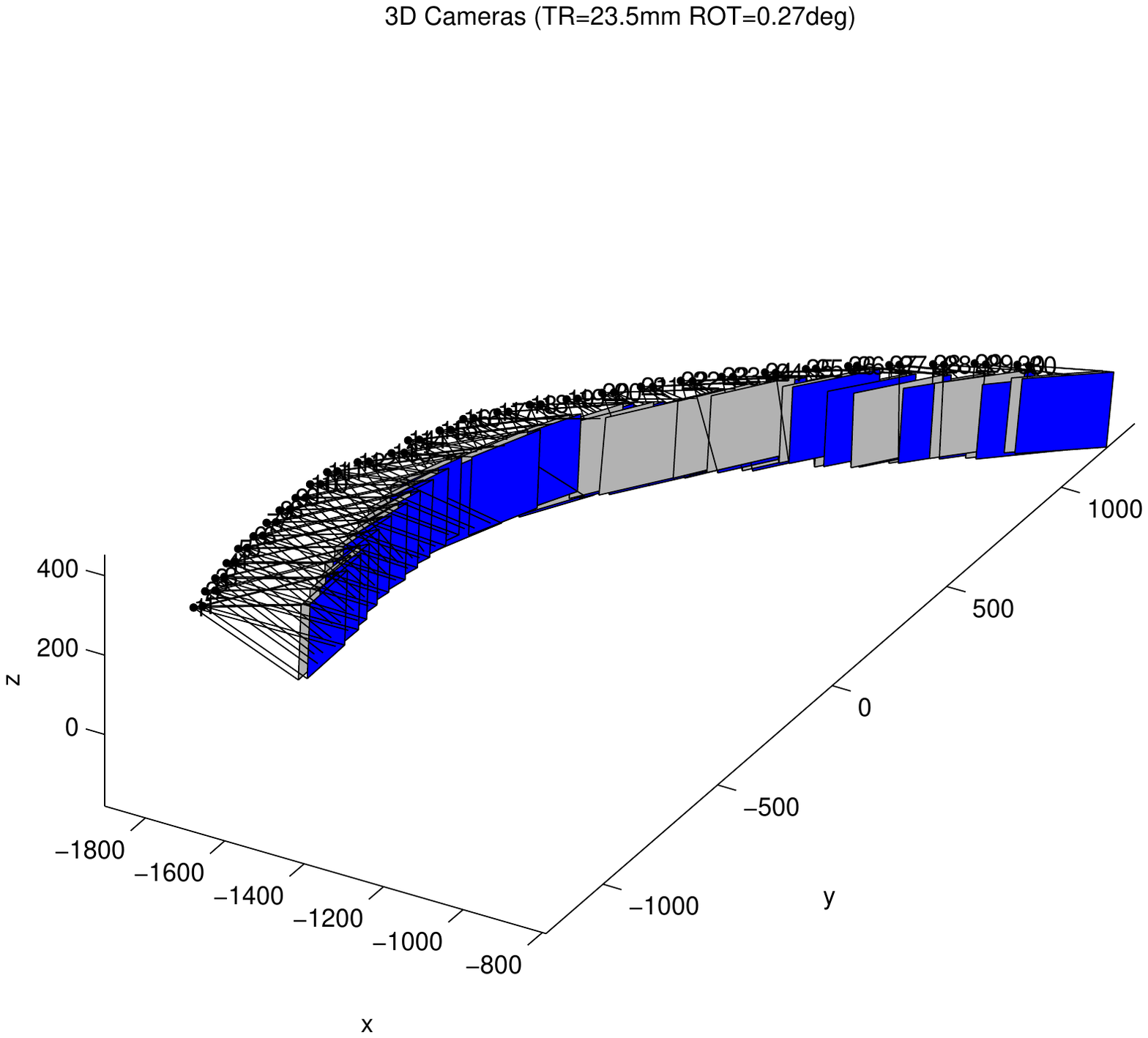} &
\includegraphics[width=0.30\textwidth]{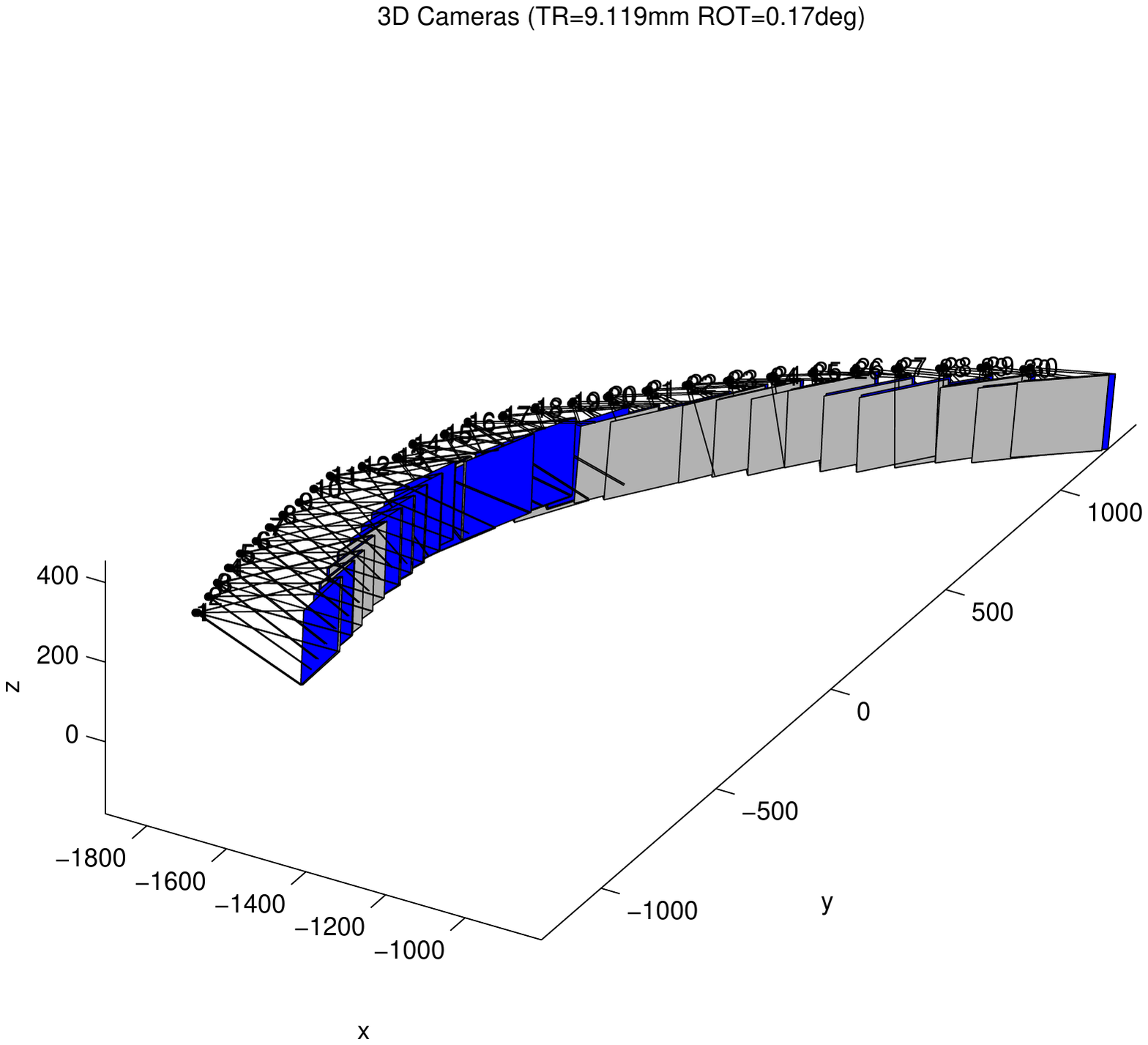} \\
\hline
\small 2D Err. = 4.53 & \small 2D Err. = 1.61 & \small 2D Err. = 0.24 \\
\hline
\end{tabular}
 \caption[camera_evolution]{Iterations of the robust perspective
   factorization algorithm in the \LineCase\ and comparison with bundle adjustment. The
   first iteration corresponds to weak-perspective factorization. The
   bundle adjustment solution is shown in grey. The bundle adjustment solution does not perform outlier treatment.}
\label{fig:camera_evolution} 
\end{figure*}

The results are summarized on
  Figures~\ref{fig:calib_results} and \ref{fig:camera_evolution} and
on Tables~\ref{tab:calib_results} and
\ref{tab:calib_results_comp}. Let us analyze in more detail these 
results. In the \CornerCase\ there are 30 cameras and 292 3-D
points. Hence, there are 8760 possible predictions out of which only
5527 are actually observed, i.e., 36\% predictions
correspond to missing 2-D data. The algorithm detected 5202 2-D inliers. An
inlier is an observation with a posterior probability greater than
$0.4$. Next, the outliers are marked as missing data. Eventually, in
this example, 285 3-D points were reconstructed (out of a total of
292) and all the cameras were correctly calibrated. The number of
iterations of the robust perspective factorization algorithm (refered
to as affine iterations) is equal to 7. On an average, there were 2.4
iterations of the EM algorithm. The obtained reconstruction
has a smaller 2-D reprojection error ($0.30$ pixels) than the one obtained by
 bundle adjustment ($0.58$ pixels).

\correction{
In any of the 3 calibration scenarios,
the proposed method outperforms bundle adjustment results, as it can
be observed in Table~\ref{tab:calib_results_comp}. This is in part due
to the fact that the bundle adjustment algorithm does not have a
mechanism for outlier rejection.
}

Figure~\ref{fig:camera_evolution} shows the evolution of the algorithm
as it iterates from a weak-perspective solution to the final
full-perspective solution. At convergence, the solution found by our
method (shown in blue or dark in the absence of colors) is practically
identical to the solution found by bundle 
adjustment (which is shown in grey). 

\begin{table*}[!htb]
\centering
\begin{tabular}{|c| l | c | c | c |}
\hline
\multicolumn{2}{|c|}{Multi-camera calibration} & \CornerCase & \LineCase & \CircleCase \\
\hline
Input&     \# Cameras & 30 & 30  & 10 \\
\cline{2-5}
&\# 3-D Points & 292 & 232  & 128 \\
\cline{2-5}
& \# 2-D Predictions & 8760  & 6960  & 1280 \\
\cline{2-5}
&\# Missing observations & 36\% & 0\%  & 33\%\\
\cline{2-5}
&\# 2-D Observations & 5527 &6960 & 863 \\
\cline{1-5}
Results &\# 2-D Inliers &5202  &6790   & 784 \\
\cline{2-5}
&\# 3-D Inliers &285  & 232  & 122 \\
\cline{2-5}
&2D error (pixels) & 0.30 & 0.19  & 0.48 \\
\cline{2-5}
&3D error (mm) & 6.91 & 2.65  & 4.57 \\
\cline{2-5}
&Rot. error (degrees) & 0.13 & 0.18  & 0.27 \\
\cline{2-5}
&Tr. error (mm) & 27.02 & 9.37  & 24.21 \\
\cline{2-5}
&\# Aff. iter. (\# EM iter.) &   7 (2.4)      &  11 (2)     &  8 (3.2)     \\
\hline
\end{tabular} 
\caption{Summary of the camera calibration results for the three setups.}
\label{tab:calib_results}
\end{table*}

\begin{table*}[!htb]
\centering
\begin{tabular}{| r | c | c | c |}
\hline
Multi-camera calibration & \CornerCase & \LineCase & \CircleCase \\
\hline
Proposed Method - 2D error (pixels) & 0.30 & 0.19  & 0.48 \\
\cline{1-4}
Bundle Adjustment - 2D error (pixels) &0.58  &0.61   & 0.95 \\
\hline
\end{tabular} 
\caption{Comparison between the proposed method and bundle adjustment
  for the three camera calibration setups.}
\label{tab:calib_results_comp}
\end{table*}

\section{3-D Reconstruction}
\label{section:3Dreconstruction}

\begin{figure*}[!htb]
\centering
\begin{tabular}{| c | r | r | r | r | r | r | r |}
\hline
 &  \multicolumn{3}{|c|}{Input Image}   & \multicolumn{2}{|c|}{3-D
   Points}   & \multicolumn{2}{|c|}{Cameras} \\ 
\hline

\begin{sideways}Middleburry Dino\end{sideways}  &
\multicolumn{3}{|c|}{ \includegraphics[width=0.23\textwidth]{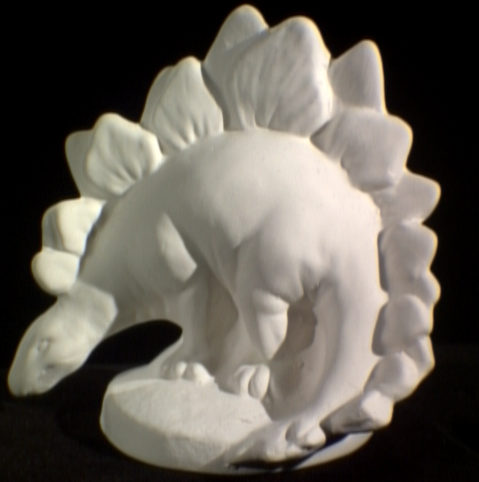} } &
\multicolumn{2}{|c|}{ \includegraphics[width=0.28\textwidth,angle=90]{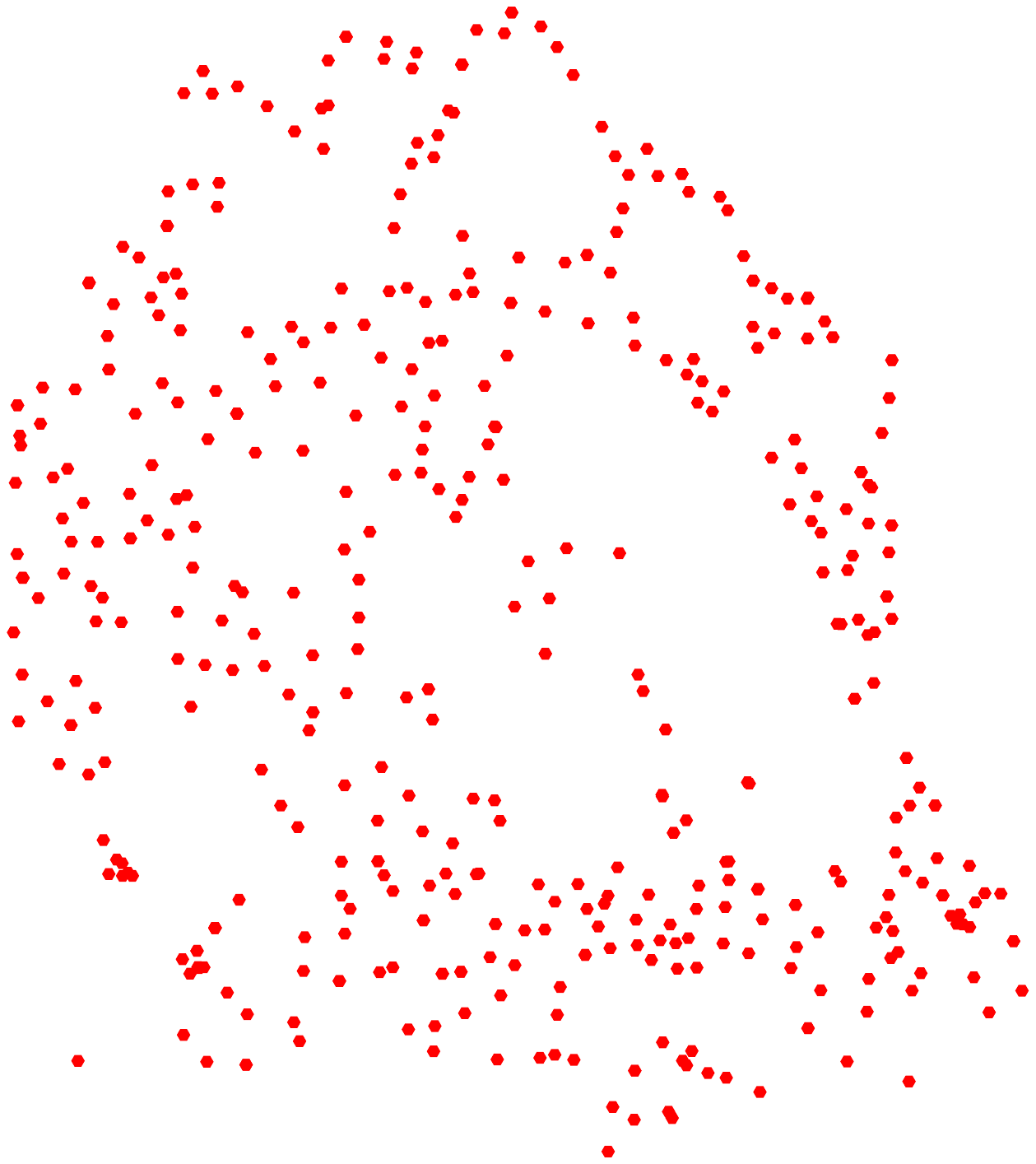} } & 
\multicolumn{2}{|c|}{ \includegraphics[width=0.28\textwidth]{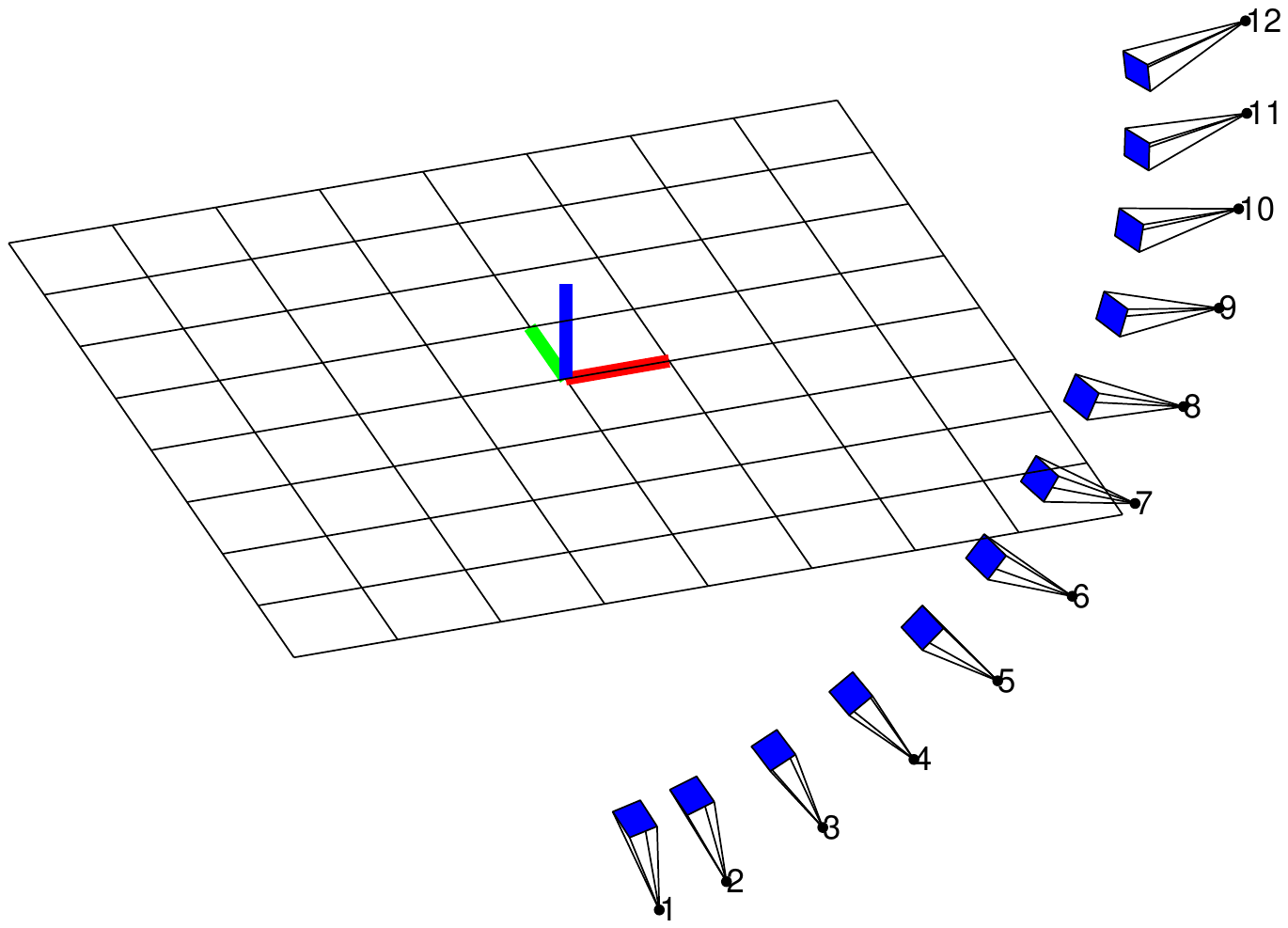} } \\ 
\hline

\begin{sideways}Middleburry Temple\end{sideways}  &
\multicolumn{3}{|c|}{\includegraphics[width=0.23\textwidth]{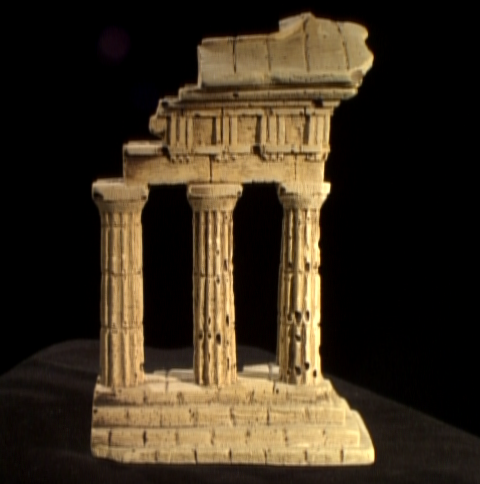}} &
\multicolumn{2}{|c|}{\includegraphics[width=0.28\textwidth,angle=90]{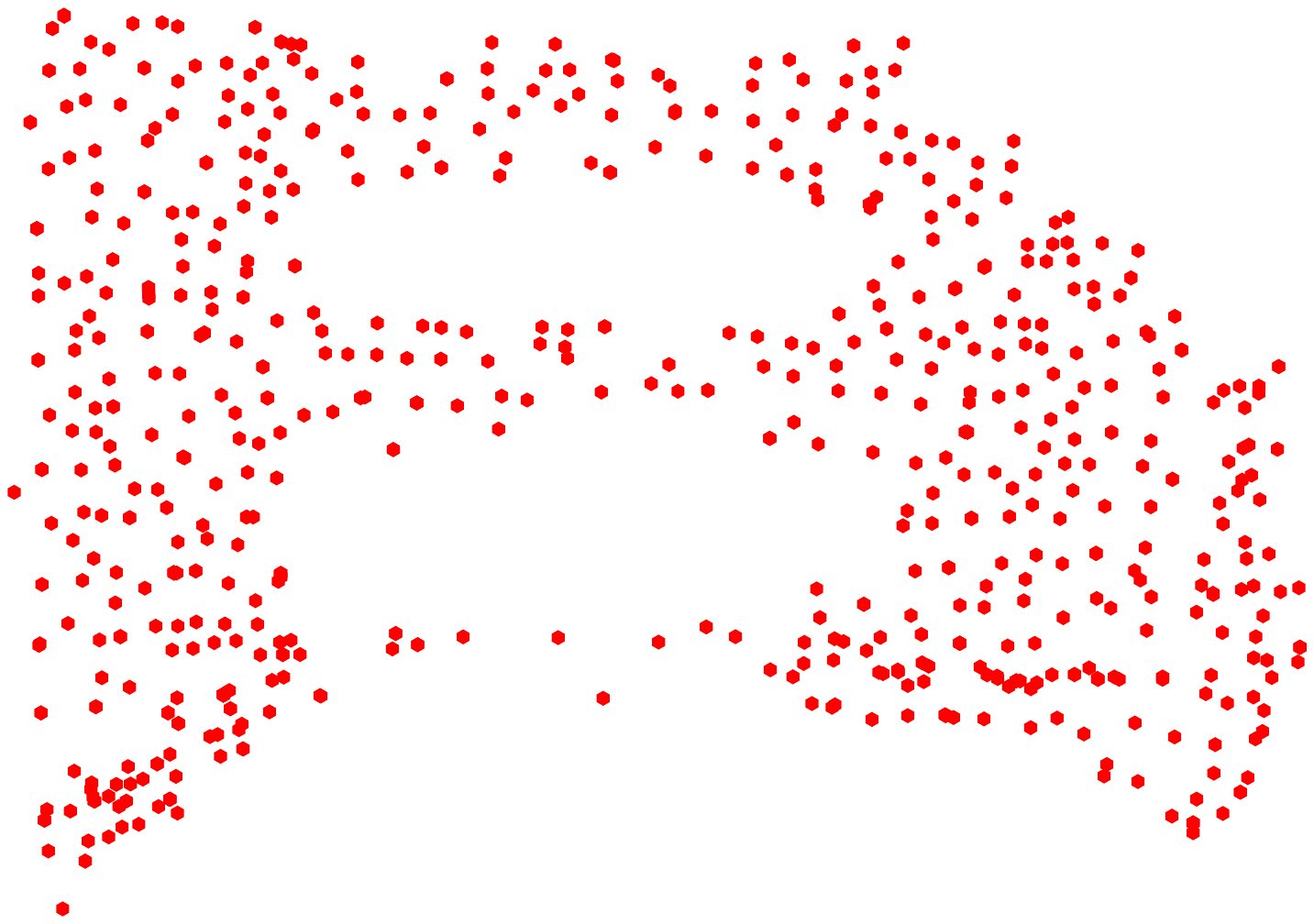}} & 
\multicolumn{2}{|c|}{\includegraphics[width=0.28\textwidth]{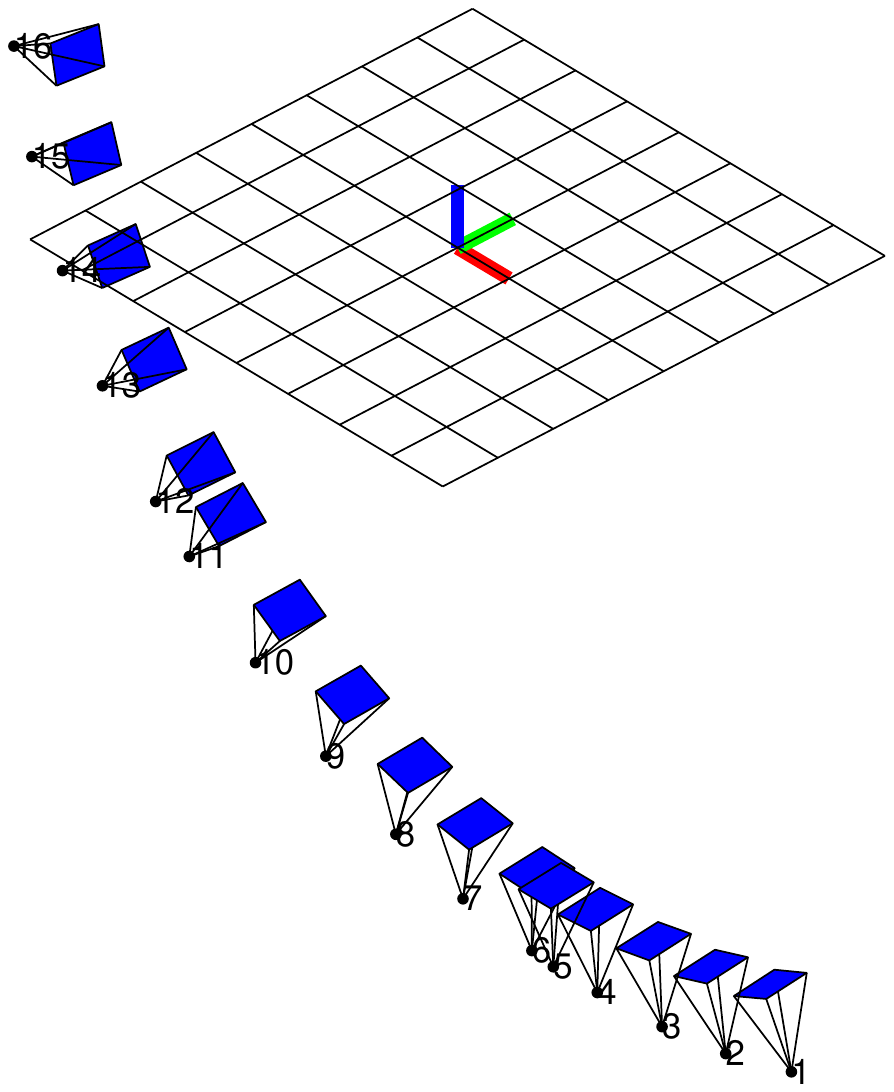}} \\ 
\hline

\begin{sideways}INRIA Box\end{sideways}  &
\multicolumn{3}{|c|}{\includegraphics[width=0.23\textwidth,width=2.0in]{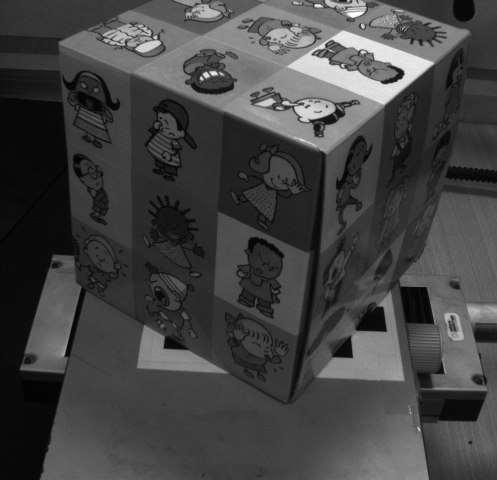}}
  &
\multicolumn{2}{|c|}{\includegraphics[width=0.28\textwidth,angle=90]{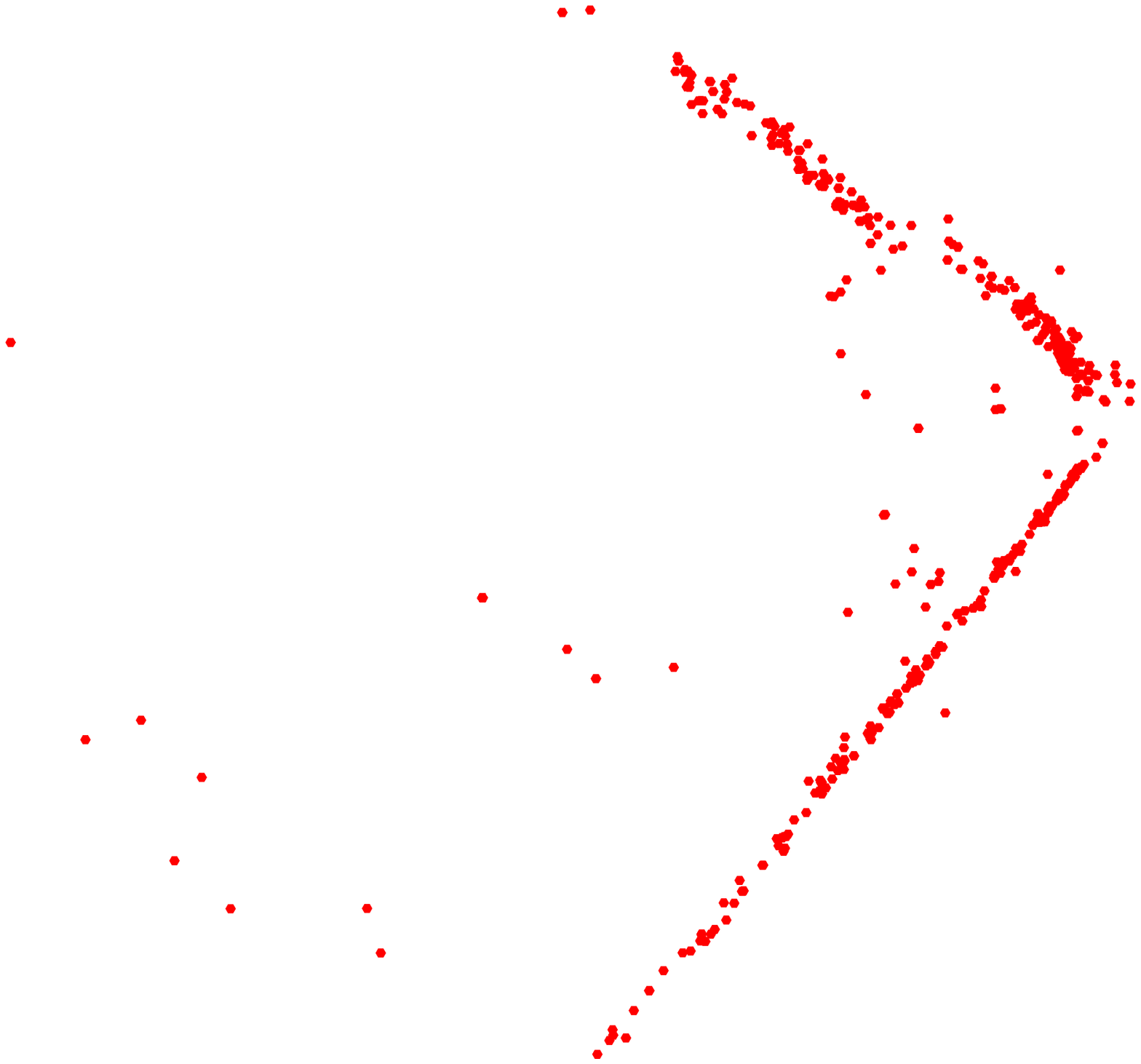}} & 
\multicolumn{2}{|c|}{\includegraphics[width=0.28\textwidth]{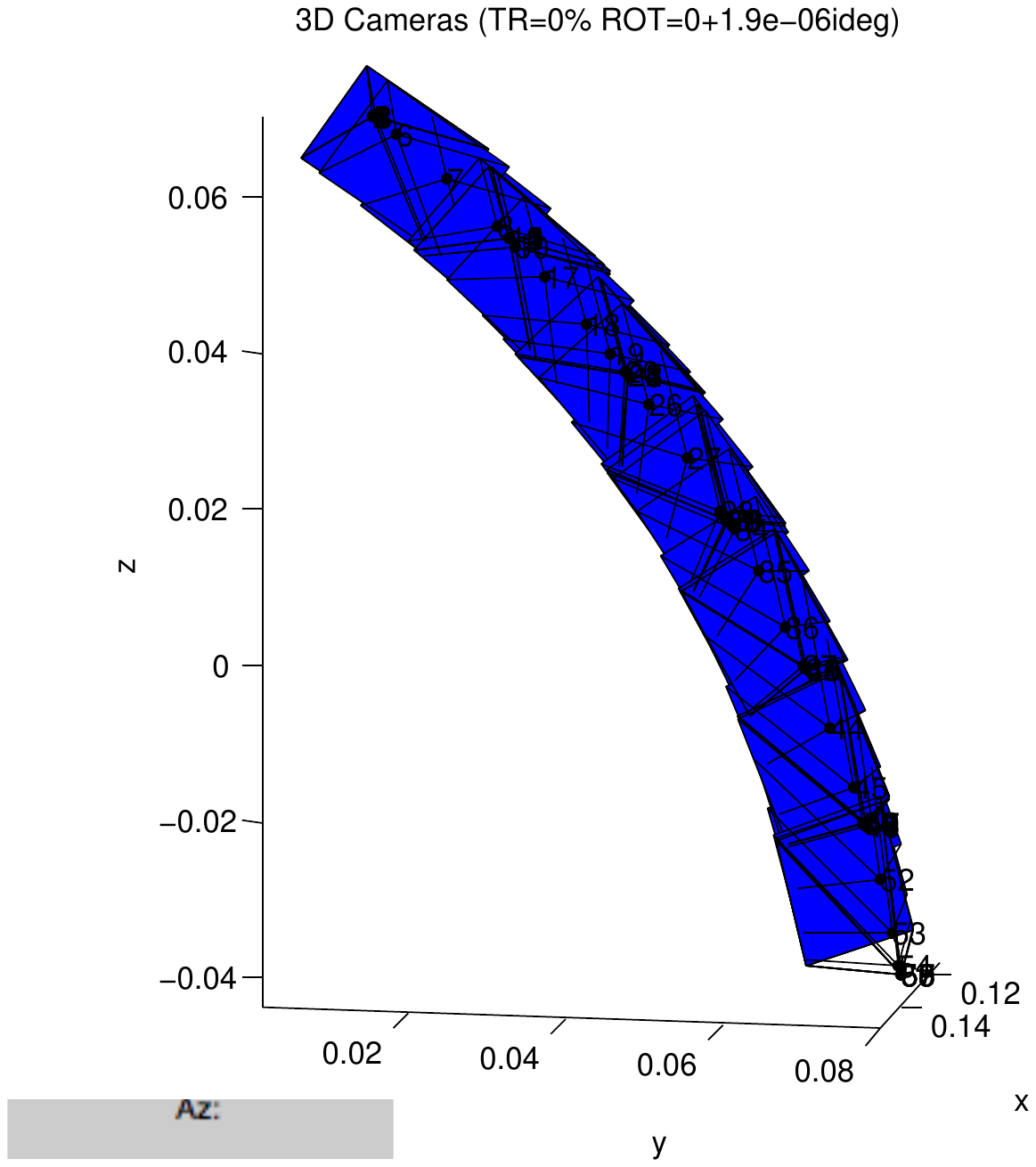}} \\ 
\hline

\end{tabular}
 
\caption[rec_results]{Ground truth data is represented in gray (light)
  colour, whereas reconstruction results are represented in blue
  (dark) colour.}
\label{fig:rec_results} 
\end{figure*}

\begin{figure*}[!htb]
\centering
\begin{tabular}{| c | r | r | r | r | r | r | r |}
\hline
 &  \multicolumn{3}{|c|}{Input Image}   & \multicolumn{2}{|c|}{3-D Points}   & \multicolumn{2}{|c|}{Cameras} \\
 \hline

\begin{sideways}Oxford Dinausor \end{sideways}  &
\multicolumn{3}{|c|}{ \includegraphics[width=0.24\textwidth]{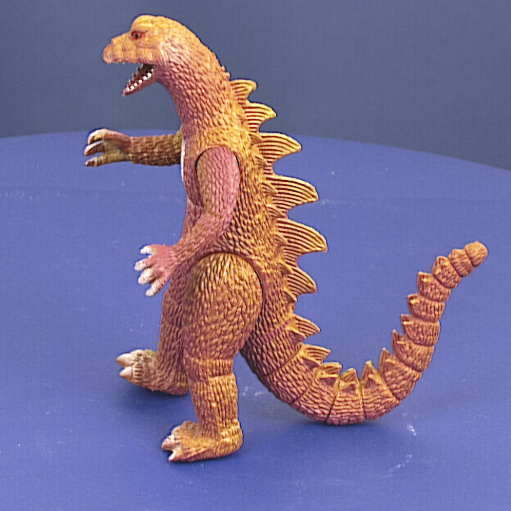} } &
\multicolumn{2}{|c|}{\includegraphics[width=0.28\textwidth]{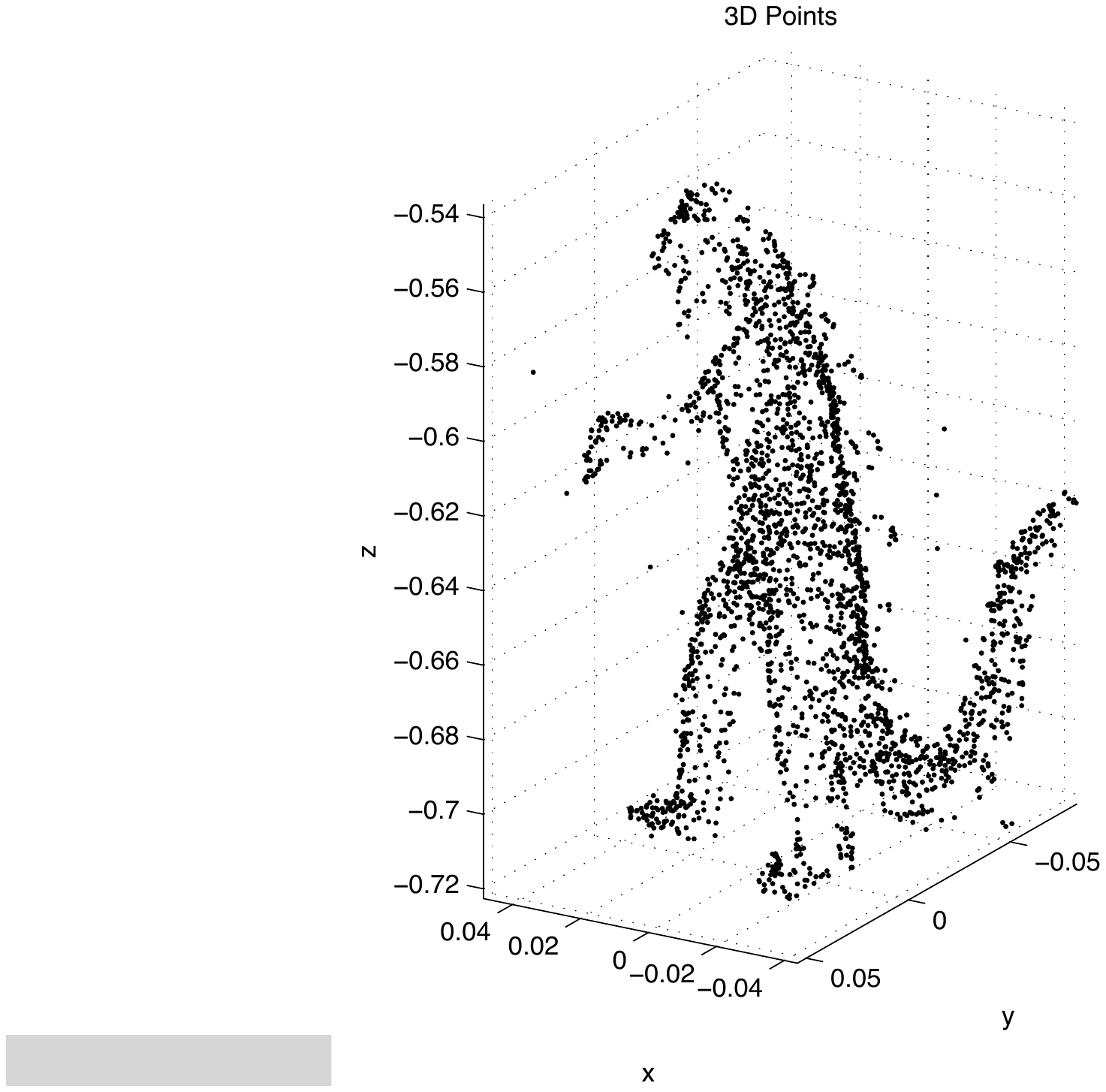}} &
\multicolumn{2}{|c|}{ \includegraphics[width=0.28\textwidth]{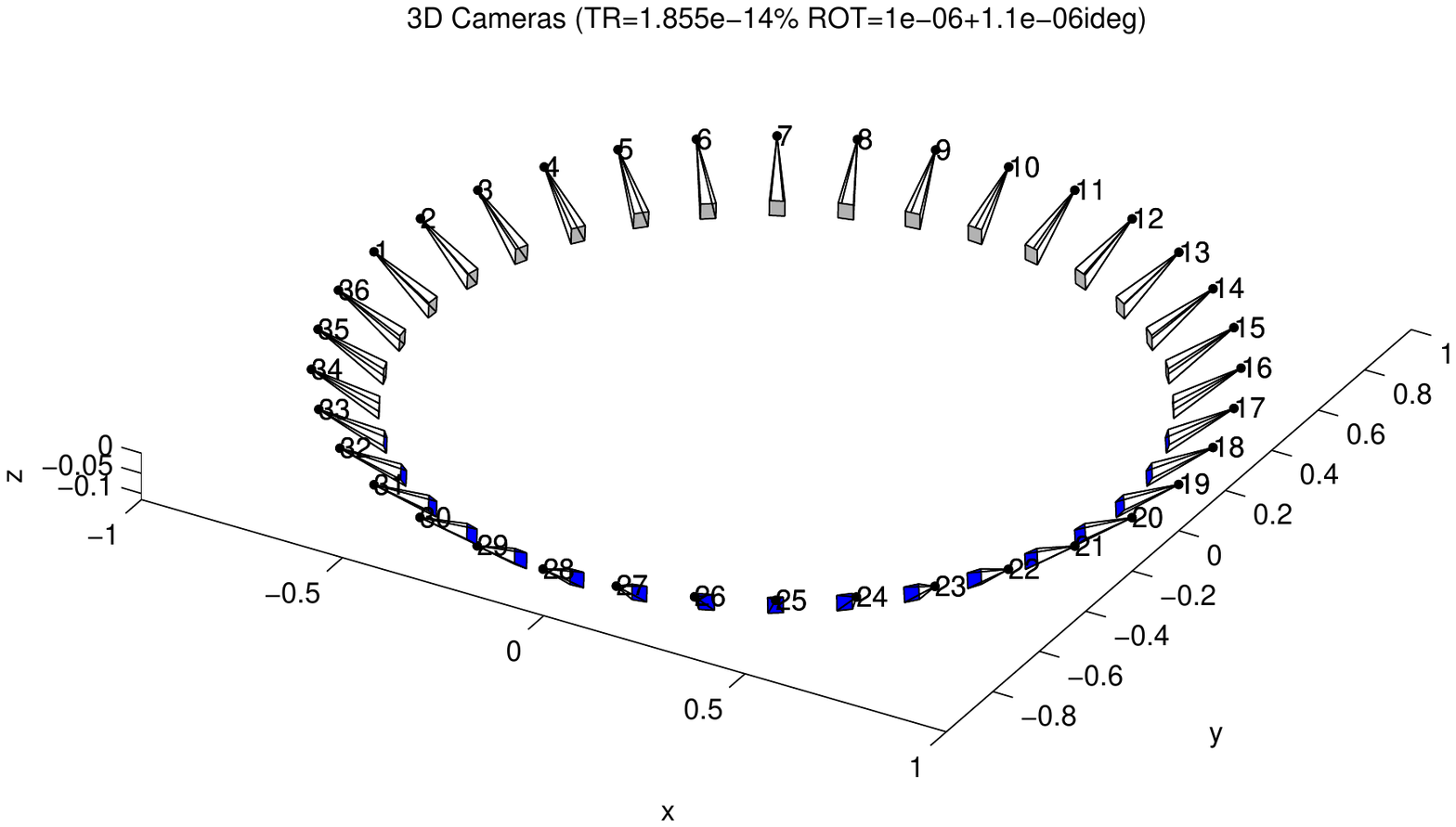} } \\ 
\hline

\end{tabular}
 
\caption[something]{Ground truth data is represented in gray (light)
  colour, whereas reconstruction results are represented in blue
  (dark) colour.}
\label{fig:rec_results_2}  
\end{figure*}

The robust perspective factorization algorithm was also applied to the
problem of 3-D reconstruction from multiple views. For this purpose we
used images of objects using a single camera and a turning table. More
specifically, we used the following data sets:
\begin{itemize}
\item The ``Dino'' and ``Temple'' data sets from the Middlebury's
  evaluation of
  Multi-View Stereo reconstruction
  algorithms;\footnote{\url{http://vision.middlebury.edu/mview/data/}}
\item The ``Oxford
  dinausor'' data set,\footnote{\url{http://www.robots.ox.ac.uk/~vgg/data/data-mview.html}} and
\item The ``Square Box'' data set.
\end{itemize}
We used the
OpenCV\footnote{\url{http://www.intel.com/technology/computing/opencv/}}
pyramidal implementation of the Lucas \& Kanade interest point
detector and tracker  \cite{Bouguet:2001} to obtain the initial set of
2-D observations. This provides the $2k\times n$ measurement matrix
$\Smat$ as well as the missing-data binary variables $\mu_{ij}$
associated with each observation. Figures~\ref{fig:rec_results} and
\ref{fig:rec_results_2}  and Table~\ref{tab:3D-results} summarize the
camera calibration and reconstruction results. For both the Middlebury
data sets (Dino and Temple) and for the Oxford data set (Dinausor) we
compared our camera calibration results with the calibration data
provided with the data sets, i.e., we measured the error in rotation
and the error in translation between our results and the data provided
in advance. 

\begin{table*}[!htb]
\centering
\begin{tabular}{|c| l | c | c | c | c |}
\hline
\multicolumn{2}{|c|}{3-D reconstruction} & Dino & Temple & Box & Dinausor \\
\hline
Input & \# Views & 12 & 16  & 64  & 36 \\
\cline{2-6}
& Size of $\Smat$ matrix & 24$\times$480 & 32$\times$758 & 128$\times$560 &
72$\times$1516\\
\cline{2-6}
& \# 2-D predictions &5760  &12128  & 35840 & 54576 \\
\cline{2-6}
& \% Missing observations &11\% &21\% &17\% & 85\% \\
\cline{2-6}
& \# 2-D Observations  & 5140 &9573   &29733 & 8331 \\
\hline
Results & \# 2-D Inliers &3124  & 6811  &25225 & 7645\\
\cline{2-6}
& \# 3-D Inliers &370 &720 &542 & 1437 \\
\cline{2-6}
& 2D error (pixels) & 0.82 & 0.93 &0.69 & 0.33 \\
\cline{2-6}
& Rot. error (degrees) & 1.49 & 2.32  &-- & 0.00 \\
\cline{2-6}
& Trans. error (mm) & 0.01 &0.01   &--  & 0.00 \\
\cline{2-6}
& \# Aff. iter. (\# EM iter.) &  7 (4)   &  7 (3.14)     & 9 (3)  & 7 (3.29)   \\
\hline
\end{tabular} 
\caption{Summary of the 3-D reconstruction results for the four data sets.}
\label{tab:3D-results}
\end{table*}

The first row in Table~\ref{tab:3D-results} introduces the test cases.
The second row corresponds to the number
of views. The third and forth rows provide the size of the measurement matrix
based on the number of views and on the maximum number of
observations over all the views. The sixth row provides the number of
actually observed 2-D points while the seventh row provides the number
of 2-D inliers (observations with a posterior probability greater than
$0.4$). The eighth row provides the number of actual 3-D reconstructed
points. One may notice that, in spite of missing data and of the
presence of outliers, the algorithm is able to reconstruct a large
percentage of the observed points. In the ``Dino'' and ``Temple''
example we compared our camera calibration results with the groundtruth 
calibration parameters provided with the Middlebury multi-stereo
dataset. Please note that these datasets are made available without 
the groundtruth 3-D data. They are typically used by the community to compare
results for 3-D dense reconstructions. The rotation error stays within 3 degrees. The translation
error is very small because, in this case we aligned the camera
centers and not the 3-D coordinates of some reconstructed points. 

The results obtained with the Oxford ``Dinausor'' need some special
comments. Because of the very large percentage of missing data, we
have been unable to initialize the solution with the
PowerFactorization method. Therefore, we provided the camera
calibration parameters for initialization. However, this kind of
problem can be overcome by using an alternative affine factorization
algorithm \cite{Tardif-al2007}.

\begin{figure*}[!htb]
\centering
\begin{tabular}{| c | c | c |  c|}
\hline
Dino & Temple & Box & Dinausor \\
\hline
\includegraphics[width=0.22\textwidth]{dino_input_image_1} &
\includegraphics[width=0.22\textwidth]{temple_input_image_1}&
\includegraphics[width=0.22\textwidth]{box_input_image_1}&
\includegraphics[width=0.22\textwidth]{dino_oxford_image_1}\\
\hline
\includegraphics[width=0.22\textwidth]{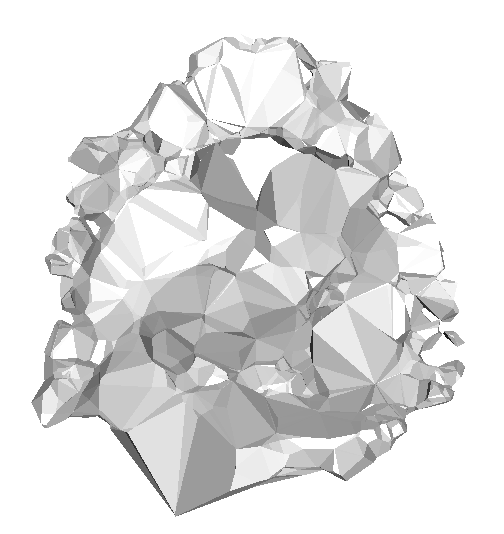} &
\includegraphics[width=0.22\textwidth]{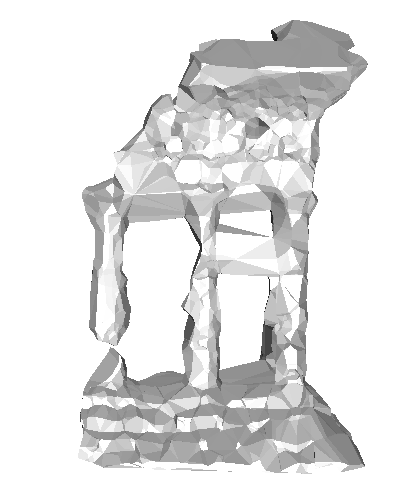} &
\includegraphics[width=0.22\textwidth]{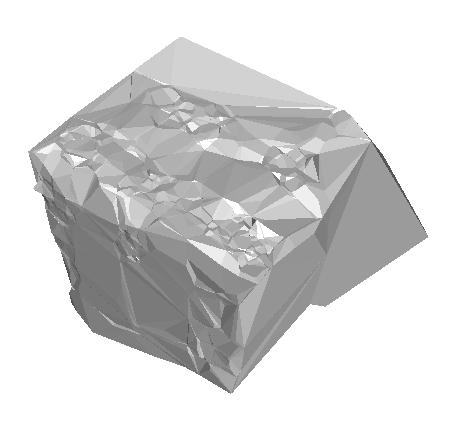} &
\includegraphics[width=0.22\textwidth]{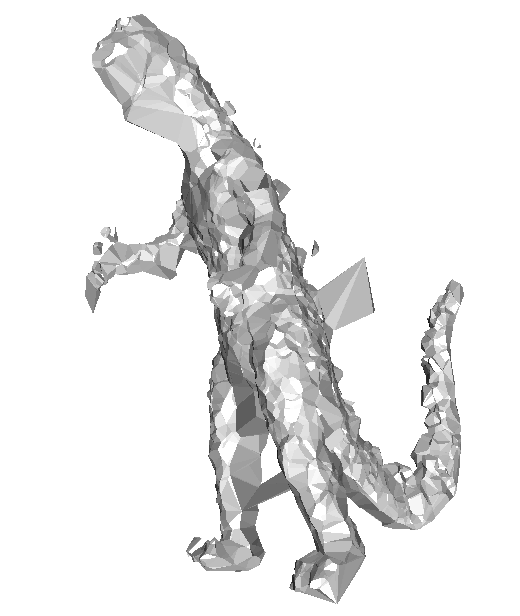} \\
\hline
\includegraphics[width=0.22\textwidth]{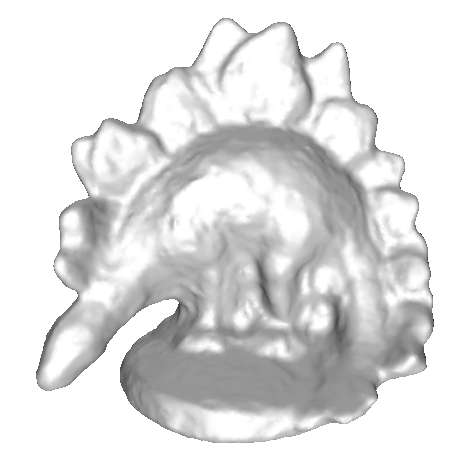} &
\includegraphics[width=0.22\textwidth]{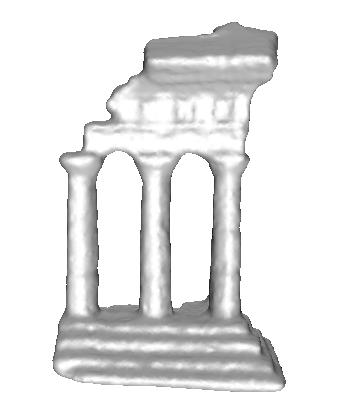} &
\includegraphics[width=0.22\textwidth]{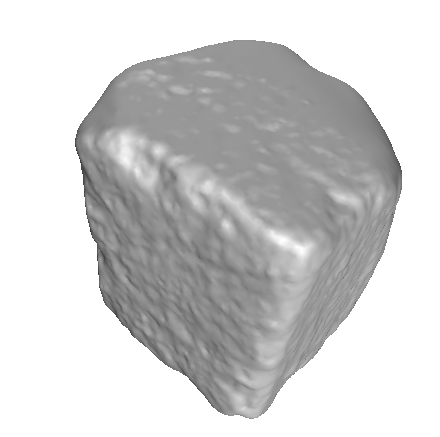} &
\includegraphics[width=0.22\textwidth]{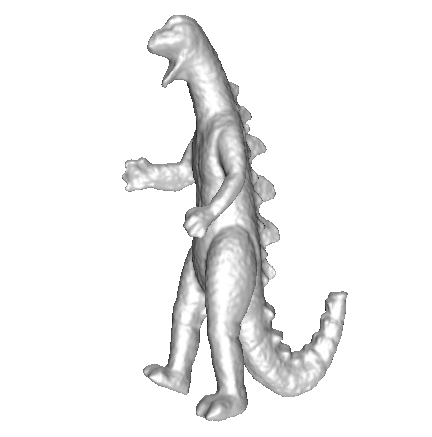} \\
\hline
\end{tabular}
\caption[rec_mesh]{Dense reconstruction results: A rough mesh obtained
from the 3-D reconstructed points (middle) and the final dense
reconstruction (bottom) after surface evolution using the method described in
\cite{ZBH07}. } 
\label{fig:rec_mesh} 
\end{figure*}

In order to further assess the quality of our results, we used the 3-D
reconstructed points to build a rough 3-D mesh and to further apply a
surface-evolution algorithm to the latter in order to obtain a more
accurate mesh-based 3-D reconstruction
\cite{ZBH07}. The results are shown on
Figure~\ref{fig:rec_mesh}.\footnote{They are also available at
\url{http://vision.middlebury.edu/mview/eval/}.}

\section{Comparison with other methods}
\label{section:EM-Mest}

As already mentioned in section~\ref{section:EMaffine}, our robust ML
estimator has strong similarities with M-estimators and their
practical implementation, i.e., IRLS \cite{Stewart:1999uq}. Previous
work on robust affine factorization has successfully used the
following reweighting function $\phi$ that corresponds to the truncated quadratic:

\begin{equation}
\label{eq:trucated_weights}
\phi(x)
=  \left\{ \begin{array}{ll} 
1 & \textrm{if $|x| < k$}\\ 
\sqrt{\frac{k^2}{x^2}} & \textrm{otherwise}\\ 
\end{array} \right. 
\end{equation}

\begin{figure}[ht]
\centering
\includegraphics[width=3in]{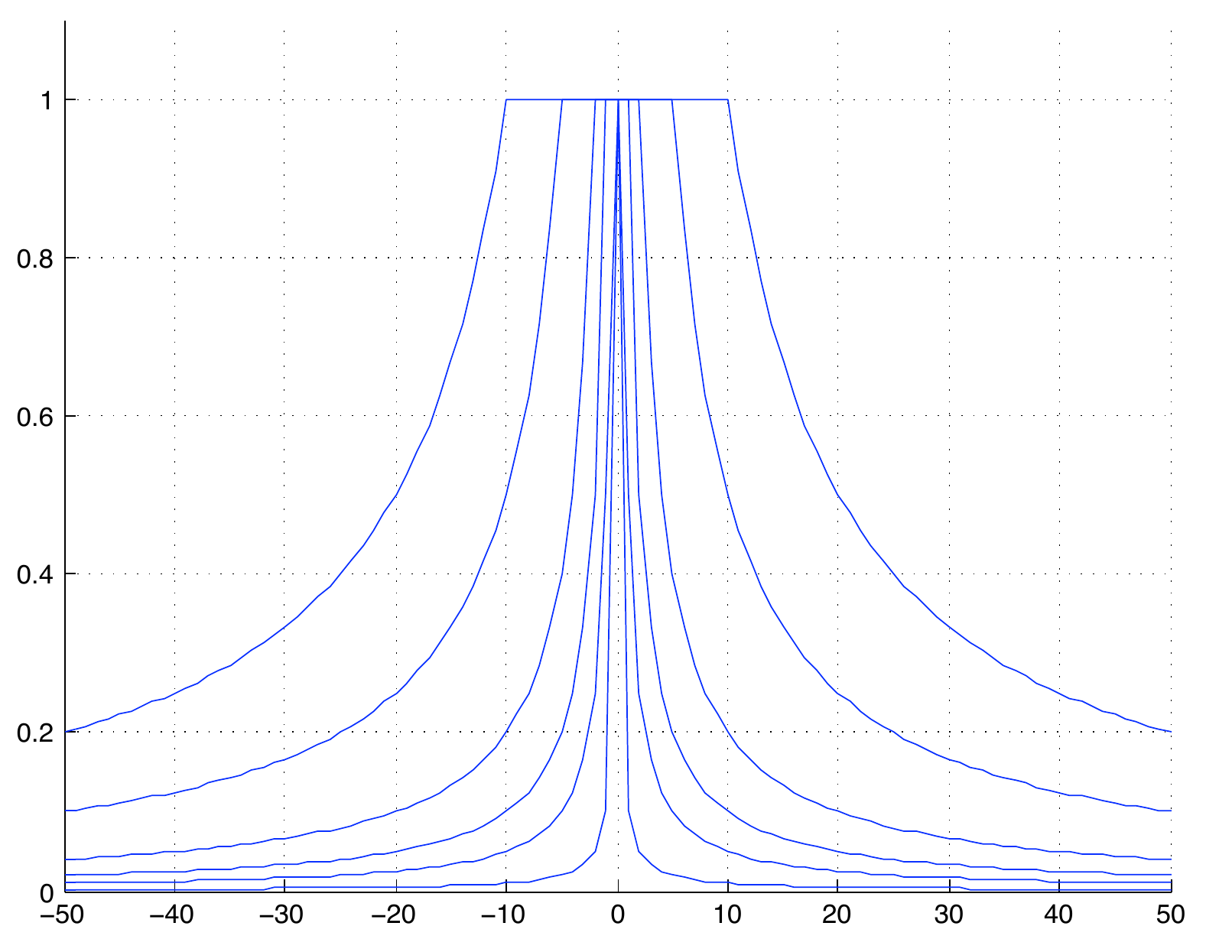}
\caption{Reweighting function for $k=0.1,0.5,1,2,5,10$.}
\label{fig:reweighting_function_comparison}
\end{figure}

It is therefore tempting to replace the EM procedure of our algorithm with
an IRLS procedure, which amounts to replace the posterior probabilities of inliers
$\alpha_{ij}^{in}$ given by eq.~(\ref{eq:inlier-posterior}) with the weights $\phi_{ij}$
given by eq.~(\ref{eq:trucated_weights}). The latter tends to zero most
quickly allowing aggressive rejection of outliers. One caveat is that
the efficiency of IRLS depends on the tuning parameter
$k$. Unfortunately the latter cannot be estimated within the
minimization process as is the case with the covariance
matrix. However, we noted that the results that we obtained do not
depend on the choice of $k$. In all the experiments reported below, we used $k=1$,
The plot of the truncated quadratic for different $k$ values is plotted in Figure~\ref{fig:reweighting_function_comparison}.

We compared the two robust methods (our EM-based robust perspective
factorization algorithm and an equivalent IRLS-based algorithm) with
five data sets for which we had the ground truth: Three multiple-camera
calibrations data sets (the \CornerCase, the \LineCase\ and the \CircleCase) and two
multi-view reconstruction data sets ({\em Dino} and {\em Temple}). 
The results of this comparison are summarized in Table~\ref{table:EM-IRLS-comparison}. 

In the {\em Corner} case the quality of the results are very similar: our
algorithm accepted 94\% of the total number of observations as inliers
and reconstructed 99.3\% of the total number of 3-D points, while IRLS
accepted all the observations as inliers and reconstructed all the 3-D
points. Similar results are obtained in the {\em Arc} and {\em Semi-Spherical} cases, where
the proposed method performs slightly better. Both algorithms were able
to reconstruct the {\em Dino} and the {\em Temple}, but our algorithm
yields more accurate results. Outlier detection is summarized in
Table~\ref{table:outliers}. 

\begin{table*}[!htb]
\begin{center}
\begin{tabular}{| c | c | c | c | c | c | c | c |}
\hline
Dataset & Method & 2-D Inliers & 3-D Inliers & 2-D err. & 3-D err. & Rot. err. & Trans. err. \\
\hline
{\em Corner} & EM & 5202 (5527) & 285 (292) & {\bf 0.30} & 6.91 & 0.13 & 27.02 \\
\cline{2-8}
& IRLS & 5526 (5527) & 288 (292) & 0.40 & 6.91 & 0.14 & 26.61 \\ 
\hline
{\em Arc} & EM & 6790 (6960) & 232 (232) & {\bf 0.19} & 2.65 & 0.18 & 9.37 \\
\cline{2-8}
& IRLS & 6960 (6960) &  232 (232) &  0.22 &  2.54 & 0.16 & 8.78 \\
\hline
{\em Semi-Spherical} & EM & 784 (863) & 122 (128) & {\bf 0.48} & 4.57 & 0.27 & 24.21 \\
\cline{2-8}
& IRLS & 862 (863) & 128 (128) & 0.62 & 4.66 & 0.29 & 23.91 \\ 
\hline
{\em Dino} & EM & 3124 (5140) & 370 (480) & {\bf 0.82} & -- & 1.49 & 0.01 \\
\cline{2-8}
& IRLS & 3411 (5140) & 390 (480) & 2.57 & -- & 2.13 & 0.01 \\
\hline
{\em Temple} & EM & 6811 (9573) & 720 (758) & {\bf 0.93} & -- & 2.32 & 0.01 \\
\cline{2-8}
& IRLS & 7795 (9573) & 731 (758) & 1.69 & -- & 2.76 & 0.03 \\
\hline
\end{tabular}
\end{center}
\caption{Comparison between robust perspective factorization results
  using EM and IRLS. The figures in paranthesis correspond to the
  total number of observations (third column) and to the total number
  of expected 3-D points (fourth column). 
}
\label{table:EM-IRLS-comparison}
\end{table*}

\begin{table}[!htb]
\begin{center}
\begin{tabular}{| c | c | c | c | c | c |}
\hline
& {\em Corner} & {\em Arc} & {\em Semi-Spherical} & {\em Dino} & {\em Temple} \\
\hline
EM &    6\%  &  2\%  &   9\%  &   39\%  &   29\%        \\
\hline
IRLS&   0\%  &  0\%  &   0\%  &   34\%  &   19\%        \\
\hline
\end{tabular}
\end{center}
\caption{Percentage of outliers detected by the two algorithms.}
\label{table:outliers}
\end{table}

\begin{figure*}[!htb]
\centering
\subfigure[\LineCase]{\includegraphics[width=0.32\textwidth]{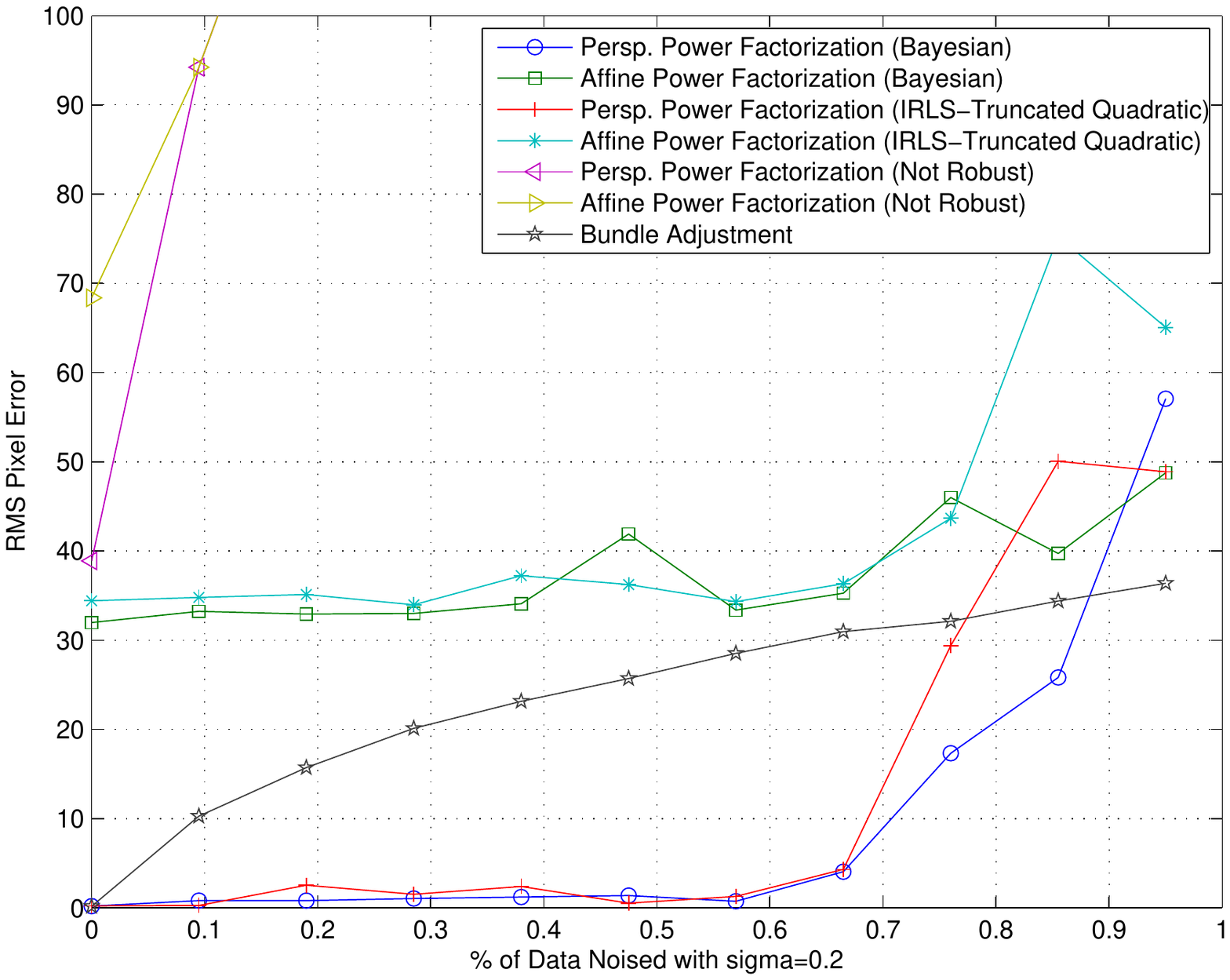} }
\subfigure[\CornerCase]{\includegraphics[width=0.32\textwidth]{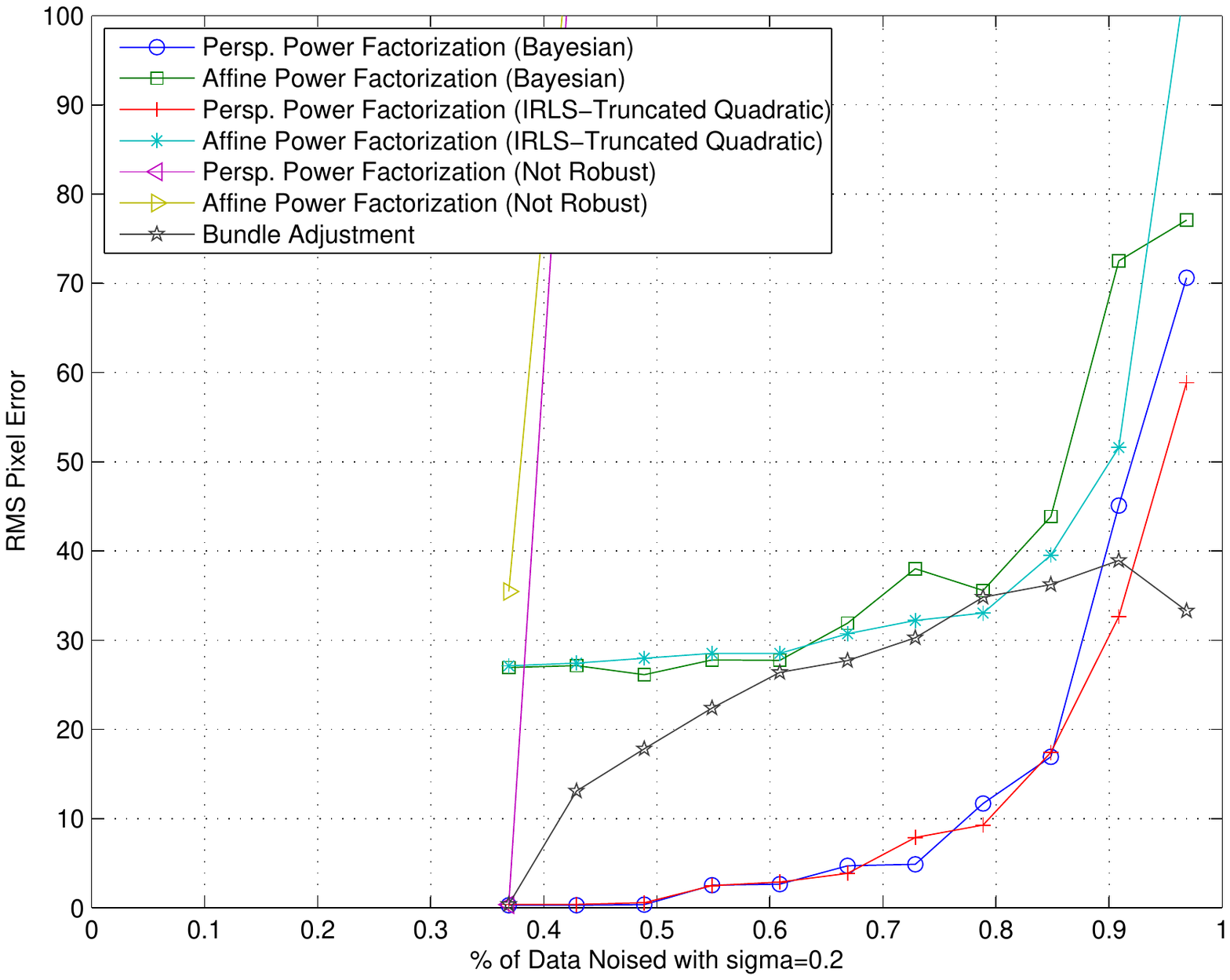} } 
\subfigure[\CircleCase]{\includegraphics[width=0.32\textwidth]{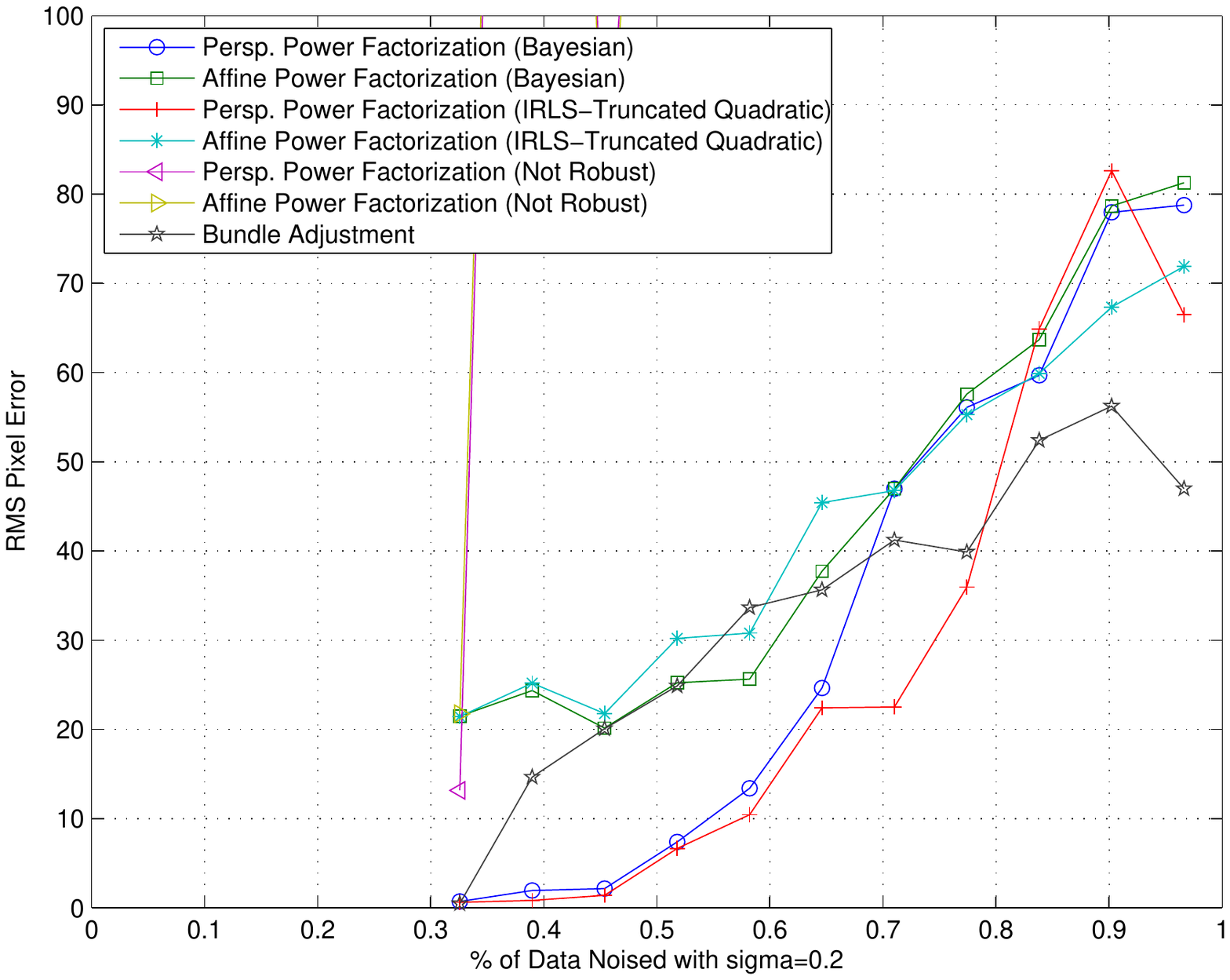} }
\caption[bundle_noise]{Behavior of various robust and non robust
  algorithms when an increasing percentage of the input data are
  corrupted by high-amplitude noise, namely $\sigma=0.2$ of the image size.}
\label{fig:alg_noise}  
\end{figure*}

A more thorough comparison with robust as well as non robust 3-D
reconstruction methods is
  provided in Figure~\ref{fig:alg_noise}. The proposed algorithm is
  denoted by "Persp. Power Factorization (Bayesian)", while the
  IRLS method is named "Persp. Power Factorization (IRLS - Truncated
  Quadratic)" and the non-robust method is called "Persp. Power
  Factorization (Not Robust)". Affine factorization algorithms
  are also presented, together with the results of bundle
  adjustment. The bundle adjustment method was always initialized
  using the PowerFactorization method.
The robust perspective factorization method proposed in this paper
  is the most resilient to high-amplitude noise. It generally performs
  better than the IRLS method and provides a clear
  advantage against the non-robust methods, which \textit{exit} the graphs as
  soon as the noise level increases. As it can be observed, in the
  \CircleCase, the solution deteriorates a lot faster in the presence
  of noise, due to the lack of the redundancy in the data (128 3-D
  points and 10 cameras, versus 292 points and 30 cameras in the
  \CornerCase\ and 232 points and 30 cameras in the \LineCase).

\begin{figure*}[!hbt]
\centering
\subfigure[Robust perspective factorization]{\includegraphics[width=0.48\textwidth]{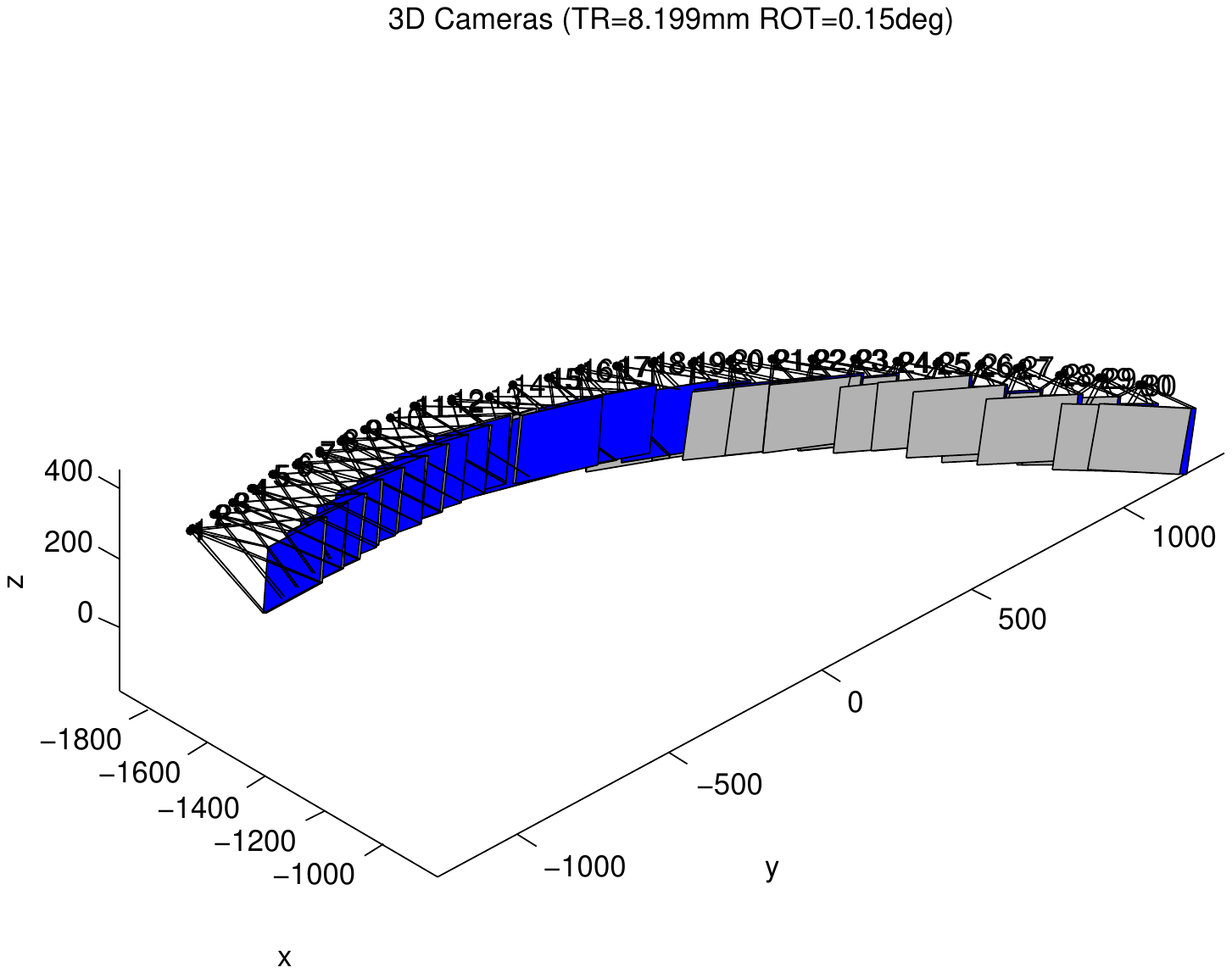} }
\subfigure[Bundle Adjustment]{\includegraphics[width=0.48\textwidth]{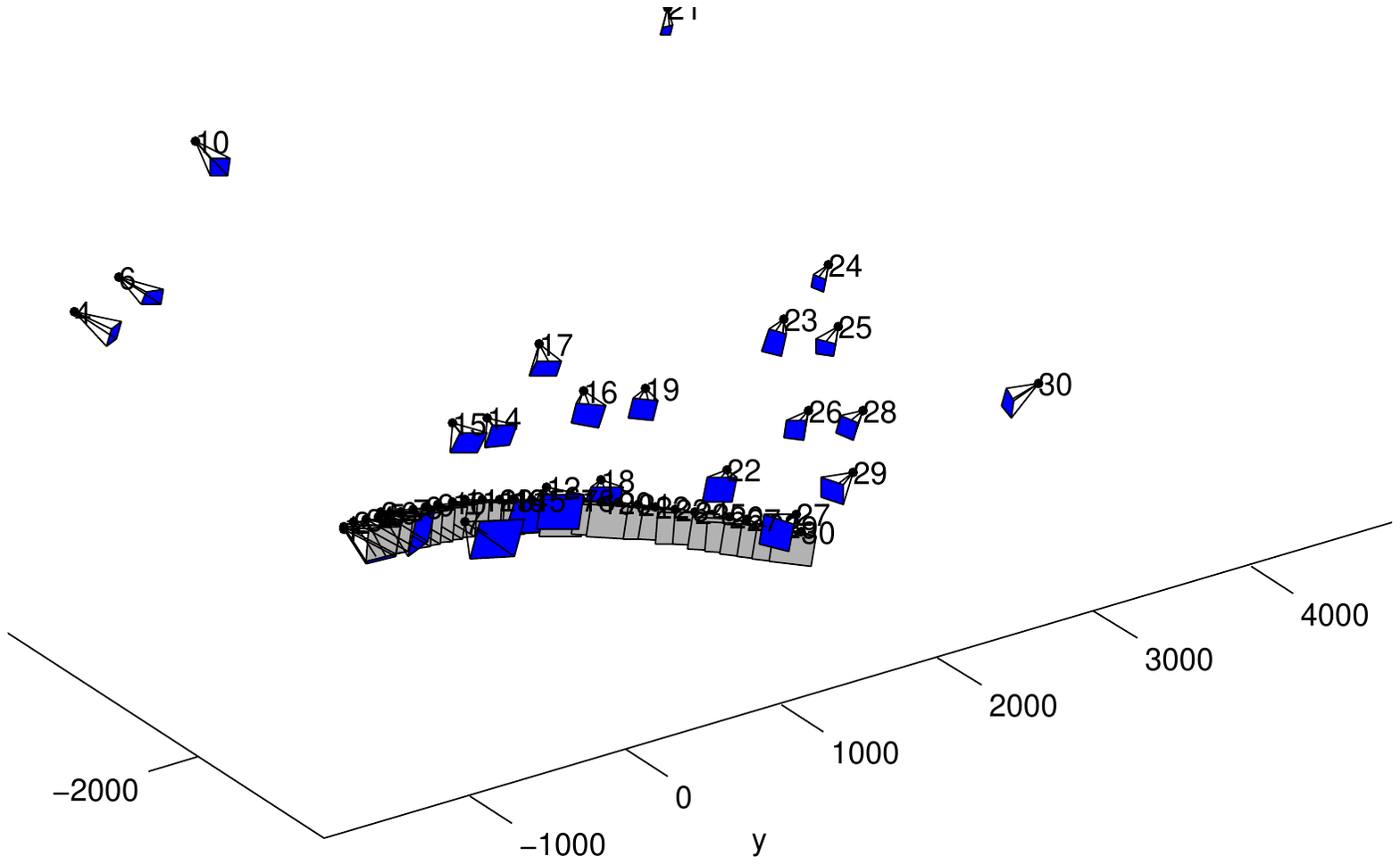} }
\caption[bundle_noise]{Calibration results in the  \LineCase\ for (a)
  the proposed method and for (b) bundle adjustment method, when 20\%
  of the input data are corrupted with high-amplitude noise, namely
  $\sigma=0.2$ of the image size. The 2-D reprojection error is of 0.71 pixels for (a) and 15.84 pixels for (b). The groundtruth is represented in gray.}
\label{fig:alg_noise_bundle}  
\end{figure*}

Figure \ref{fig:alg_noise_bundle} compares our method (a), with the
bundle adjustment method (b), in the  \LineCase and when $20\%$ of the input data was
corrupted by high-amplitude noise ($\sigma=0.20$ of the image
size). On both figures the ground truth is shown in grey and the
result of the algorithm is shown in blue (or dark in the absence of
color). Notice that with this level of data perturbation, bundle
adjustment completely failed to find the correct solution.


\section{Conclusions}
\label{section:conclusions}
In this paper we described a robust factorization method based on
data clustering and on the EM algorithm. First we recalled the
classical maximum-likelihood approach within which all the observations are
supposed to be independent and identically distributed. This a\-mounts
to classify all the observations in one cluster -- inliers. Next we
considered a mixture model within which the likelihood of the inlier
class has a normal distribution and the likelihood of the outlier
class has a uniform distribution. This naturally leads to ML with
missing variables which is solved in practice via the
Expectation-Maximization algorithm. We formally derived the latter in
the specific case of 3-D reconstruction and of a Gaussian/uniform
mixture; This allowed us to
rely on EM's convergence properties.

Moreover, we devised two shape and motion algorithms: (i)~affine
factorization with EM and (ii)~robust perspective factorization, the
former residing in the inner loop of the latter. These two algorithms
are very general since they can accomodate with any affine
factorization and with any iterative perspective
factorization methods.

We performed
extensive experiments with two types of data sets: multiple-camera
calibration and 3-D reconstruction. We compared the calibration
results of our
algorithm with the results obtained using other methods such as the bundle adjustment
technique and IRLS. It is interesting to notice that there is almost no
noticeable quantitative difference between our algorithm and a
non-linear optimization method such as bundle adjustment. The 3-D
reconstruction results obtained with a single camera and objects
lying on a turntable are also very good. Whenever possible, we
compared our results with ground-truth data, such as the
external camera parameters provided by the Middlebury multi-view
stereo data set. In order to further assess the 3-D reconstruction
results, we used the output of the robust perspective factorization
method, namely a cloud of 3-D points, as input of a mesh-based
reconstruction technique. 

Our Gaussian/uniform mixture model and its associated EM algorithm may
well be viewed as a robust regression method in the spirit of
M-estimators. We compared our method with IRLS using a truncated
quadratic loss function. The results show that our method performs slightly
better, although we believe that these results are only
preliminary. A thorough comparison between outlier detection using
probability distribution mixture models on one side, and robust loss
functions on the other side is a topic in its own right. In the future
we plan to extend our method to deal the more difficult problem of
multiple-body factorization.

\appendix

\newcommand{\sterm}{(\svect_{ij} -\hat{\svect}_{ij}(\vv{\theta}))}

\section{Derivation of equation (\ref{eq:covariance-ML})}
\label{appendix:A}

We recall eq.~(\ref{eq:minimization-inliers}):
\begin{equation}
Q_{ML}=
\frac{1}{2}
\sum_{i,j} \bigg(
\sterm\tp\Cmat\inverse\sterm + \log (\det\Cmat) 
\bigg) \nonumber
\end{equation}

Taking the derivative with respect to the entries of the 2$\times$2
matrix $\Cmat$ we obtain:

\begin{align}
\frac{\partial Q_{ML}} {\partial \Cmat} &=
\frac{1}{2} \sum_{i,j}
\bigg( 
- \Cmat^{-\tp}\sterm\sterm\tp\Cmat^{-\tp} + \Cmat^{-\tp}
\bigg) \\
&= - \Cmat^{-\tp} \bigg( \frac{1}{2} \sum_{i,j} \sterm\sterm\tp \bigg) \Cmat^{-\tp} + \frac{m}{2} \Cmat^{-\tp}
\end{align}

where $m=k \times n$. By setting the derivative to zero we obtain eq.~(\ref{eq:covariance-ML}):



\begin{equation}
\Cmat = \frac{1}{m} \sum_{i,j}
\sterm\sterm\tp
\nonumber
\end{equation}

\section{Derivation of equation (\ref{eq:ML-sigma})}
\label{appendix:B}

When considering isotropic covariance, $\Cmat = \sigma^2\Imat$, hence
$\det\Cmat=\sigma^4$, and the equation becomes: 

\begin{equation}
Q_{ML}=
\frac{1}{2}
\sum_{i,j} \bigg(
\frac{1}{\sigma^2} \|\svect_{ij} - \hat{\svect}_{ij}(\vv{\theta})\|^2
+ 2\log(\sigma^2) 
\bigg)
\end{equation}

By taking the derivative with respect to $\sigma^2$, we obtain:

\begin{align}
\frac{\partial Q_{ML}} {\partial \sigma^2} &=
\frac{1}{2}
\sum_{i,j}
\bigg(
-\frac{1}{(\sigma^2)^2} \|\svect_{ij} - \hat{\svect}_{ij}(\vv{\theta})\|^2 + 2\frac{1}{\sigma^2}
\bigg) \\
&= \frac{1}{2}
\sum_{i,j} 
\bigg(
\frac{2\sigma^2 - \|\svect_{ij} - \hat{\svect}_{ij}(\vv{\theta})\|^2}{\sigma^2}
\bigg)
\end{align}

By setting the derivative to zero, we obtain eq.~(\ref{eq:ML-sigma}):



\begin{equation}
{\sigma^2} = \frac{1}{2m} \sum_{i,j} \alpha_{ij}^{in}
\| \svect_{ij} -
\hat{\svect}_{ij}(\vv{\theta}^{\ast}))\|^2 \nonumber
\end{equation}

\bibliographystyle{elsarticle-num}   


\end{document}